\documentclass{article} 
\usepackage{iclr2027_conference,times}

\usepackage{amsmath,amsfonts,bm}

\def\eqref#1{equation~\ref{#1}}

\def\1{\bm{1}}

\DeclareMathAlphabet{\mathsfit}{\encodingdefault}{\sfdefault}{m}{sl}
\SetMathAlphabet{\mathsfit}{bold}{\encodingdefault}{\sfdefault}{bx}{n}

\DeclareMathOperator*{\argmax}{arg\,max}
\DeclareMathOperator*{\argmin}{arg\,min}

\usepackage{hyperref}
\usepackage{url}
\usepackage{graphicx}
\usepackage{subcaption}
\usepackage{caption}
\usepackage{mathtools}
\usepackage{cleveref}
\usepackage[dvipsnames,svgnames,x11names]{xcolor}

\newcounter{daggerfootnote}
\newcommand*{\daggerfootnote}[1]{%
    \setcounter{daggerfootnote}{\value{footnote}}%
    \renewcommand*{\thefootnote}{\fnsymbol{footnote}}%
    \footnote[2]{#1}%
    \setcounter{footnote}{\value{daggerfootnote}}%
    \renewcommand*{\thefootnote}{\arabic{footnote}}%
}

\newcommand{\cA}{\mathcal A}
\newcommand{\bx}{\mathbf{x}}
\newcommand{\ba}{\mathbf{a}}
\newcommand{\bh}{\mathbf{h}}
\newcommand{\bH}{\mathbf{H}}
\newcommand{\bu}{\mathbf{u}}
\newcommand{\bd}{\mathbf{d}}

\newcommand{\bbR}{\mathbb{R}}

\definecolor{wzpurple}{HTML}{7A3E9D}

\title{A  theory of platonic representations in language models }

\author{Darshil Doshi\thanks{Co-first authors. Correspondence to: ddoshi3@jh.edu, mwyart1@jh.edu} \\
Johns Hopkins University
\And
Wenjie Zhou\footnotemark[1] \\
Johns Hopkins Univeristy
\And
Corinna Elena Wegner \\
EPFL
\And
Daniel J. Korchinski \\
EPFL \\
\And
Santiago Acevedo \\
SISSA
\And
Matthieu Wyart\\
\makebox[\linewidth][c]{Johns Hopkins University \& EPFL}
}

\iclrfinalcopy 
\begin{document}

\maketitle

\begin{abstract}
Representations of translated sentences are similar in the inner layers of
multilingual language models -- an observation connected to the platonic
representation hypothesis, yet unexplained theoretically. We provide an
explanation based on the assumption that data have a hidden hierarchical
structure whose abstract levels are shared across languages while surface
levels are modality- or language-specific. Concretely, we generate synthetic languages
from probabilistic context-free grammars  sharing upper-level but not lower-level production rules.
In this setting the Bayes-optimal next-token predictor is belief propagation (BP);
encoding its messages in successive layers yields analytical predictions
that agree well with transformers trained on the same data. The framework
explains why cross-lingual similarity peaks in middle layers, coexists with
language-specific structure, and strengthens with language proximity, model
quality and data exposure. It distinguishes similarity (shared neighborhood geometry) 
from alignment (shared coordinates), showing that the latter occurs when
code-switched data, i.e. mixed-language sentences, are abundant enough. 
It further predicts that subtracting from each layer the component linearly predictable from the preceding one increases cross-lingual similarity, which we confirm in pretrained LLMs.\daggerfootnote{Code to reproduce our results: \url{https://github.com/d-doshi/platonic.git}}

\end{abstract}


\section{Introduction}
The platonic representation hypothesis posits that models trained on vastly
different data converge toward a shared representation of reality
\citep{huh2024platonic}. Consistent with this view, semantically similar
inputs elicit similar representations in the inner layers of trained networks
\citep{acevedo2026quantitative}. Remarkably, sentences and their translations
are similarly represented inside multilingual language models, and the same
holds across modalities: if two pictures are encoded similarly by a vision
transformer, so are their captions by a language model
\citep{huh2024platonic}. 
Here, similarity of representations means that they tend to share a local geometry or neighborhood structure.
Despite the considerable attention
these results have attracted, a theoretical framework explaining these
phenomena is missing. 
A satisfactory theory should account for the following observations:

\begin{figure}[!t]
    \centering
    \includegraphics[width=\linewidth]{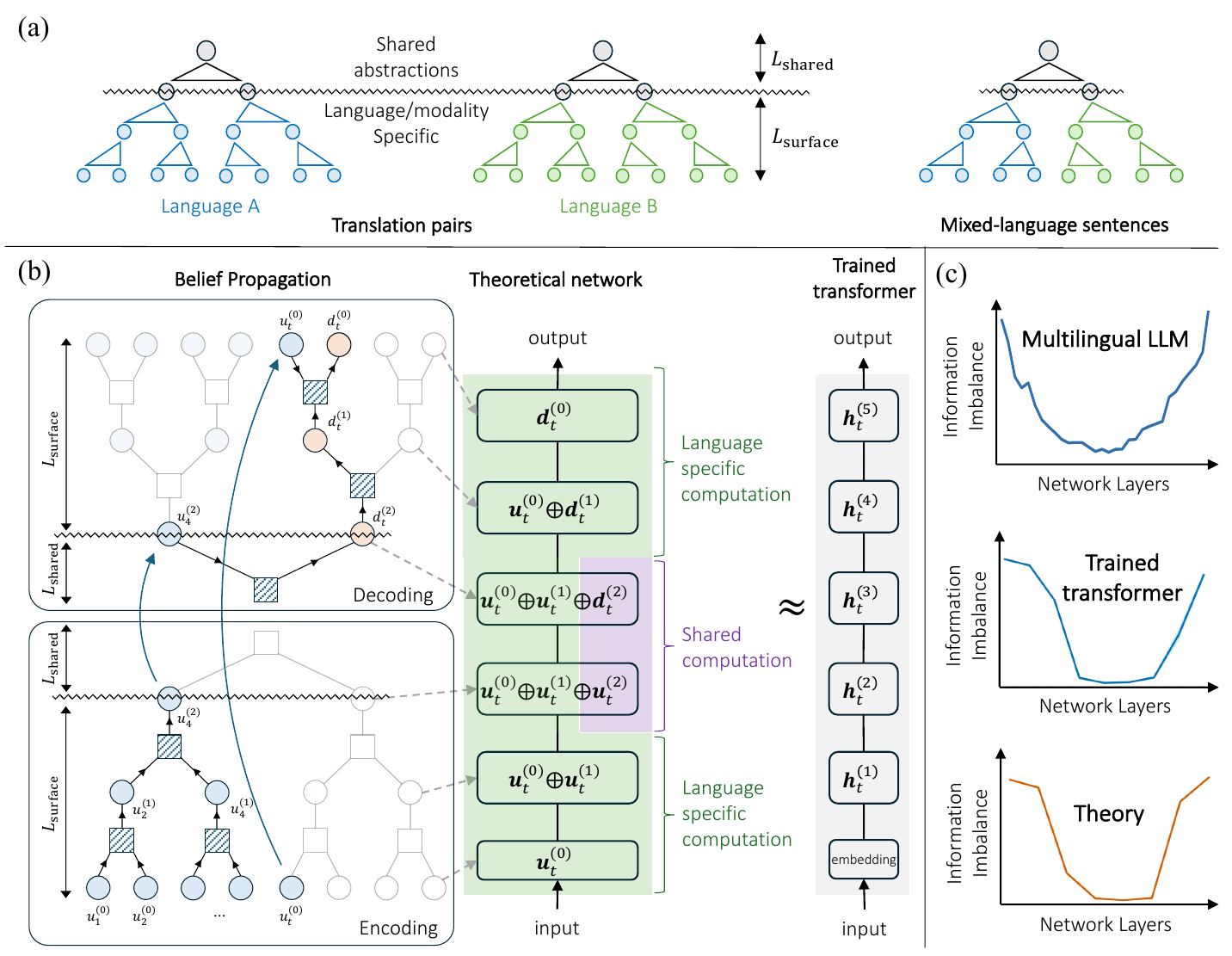}
    \caption{\textbf{(a)} Multilingual Random Hierarchy Model. Upper-level production rules are shared between languages, while lower-level rules
    are language-specific. Zigzag line denotes this separation. Mixed sentences are generated by replacing
    language-specific subtrees.
    \textbf{(b)} Optimal next-token prediction via inference with belief
    propagation (BP). \textit{Encoding:} From $t$ observed tokens, beliefs on higher-and-higher latent variables are sequentially computed and propagated upward. The highlighted nodes denote the encoded beliefs, and the lined boxes denote the computations. \textit{Decoding:} Posterior probabilities of latent variables are computed iteratively using the encoded information, toward lower-and-lower levels,  ultimately leading to the probability of the $(t+1)$-th token. From this exact algorithm, a theoretical network is constructed, whose layerwise representations encode the steps of the BP algorithm. The computations below $L_{\mathrm{surface}}$ are language-specific, while those above $L_{\mathrm{surface}}$ have shared components. This leads to a maximal  cross-lingual similarity near the center of theoretical networks, confirmed in trained transformers. See Sec.\ref{sec:bp} for details. 
    \textbf{(c)} Cross-lingual similarity measured with Information
    Imbalance (where lower Information Imbalance corresponds to higher similarities, see Appendix~\ref{secapp:similarity_measures}). OLMo-3-7B tested on English and Polish \citep{olmo2026olmo3}, transformers trained on multilingual RHM
    data, and theoretical networks constructed with BP -- all exhibit similar
    II profiles, with strongest similarity in the middle layers. }
    \vspace{-1em}
    \label{fig:intro}
\end{figure}

1. \emph{Similarity coexists with language-specific structure.}
Multilingual language models do not collapse their representations into a
fully neutral interlingua: they retain language-specific information
alongside shared conceptual structure
\citep{singh2019bert, chang2022geometry, dumas2025separating}.

2. \emph{Similarity is organized systematically across layers.} In
multilingual LLMs, representations of translated sentences are most similar
in the middle layers, surrounded by more language specific processing near
the input and output
\citep{wendler2024llamas, zhao2024how, liu2025middle, dumas2025separating,
inaba2025bilingual,huh2024platonic}.

3. \emph{Similarity increases with language proximity.} Cross-lingual
similarity is stronger between structurally and typologically close
languages \citep{jones2021massively, peng2024concept}.

4. \emph{Similarity tracks model quality and data exposure.} It
strengthens as models acquire better representations through training and
increased capacity \citep{peng2024concept, inaba2025bilingual}, and it is
weaker for low-resource languages, which exhibit poorer cross-lingual
transfer \citep{wu2020languages, miao2024enhancing}.


5. \emph{Beyond similarity, models can achieve alignment.} Models
fine-tuned on a downstream task in one language often transfer zero-shot to
others \citep{pires2019multilingual}. This reveals a stronger property than geometric
similarity, which we call \emph{alignment}: the two languages encode shared
information along the same directions, so that a readout learned in one
operates in the other without refitting -- a property that
rotation-invariant similarity measures cannot detect. Mixed-language
contexts, including synthetic code-switching during pretraining, reinforce
this alignment \citep{wang2025investigating}.


To explain these phenomena, we introduce a framework based on the assumption that language and
images have a hidden hierarchical structure -- a central notion in both
linguistics \citep{chomsky1956three} and computer vision
\citep{mumford2010pattern, grenander2007pattern} -- and that the highest,
most abstract levels of this hierarchy are shared across languages and
modalities, while the lowest levels encode surface aspects (vocabulary,
low-level syntax) that differ between them, as illustrated in
Fig.~\ref{fig:intro}(a). A framework considering generative models  trained on such data accounts for all the facts listed
above \footnote{Extending this framework to representation learning algorithms such as joint embedding methods is left for the future.}. It also makes a new prediction:
subtracting from each hidden layer the component linearly predictable from
the previous layer isolates the information newly computed at that depth,
and this residual is more suitable for predicting the neighbor structure of  other languages  than the raw
representations. We confirm this prediction both in our controlled setting
and in pretrained LLMs processing natural languages.

In practice, we design synthetic languages within the class of context-free
grammars \citep{chomsky1956three}. Specifically, we consider a multilingual extension of the Random
Hierarchy Model \citep{cagnetta2024towards} in which languages share the upper production rules but
differ in the lower ones. The model yields two kinds of predictions:
computational, by training GPT-2-style transformers on these data for
next-token prediction; and analytical, from belief propagation -- the
algorithm that solves next-token prediction optimally on such data, sketched
in Fig.~\ref{fig:intro}(b). Belief propagation is iterative, and assuming
that its successive messages are linearly encoded in successive layers turns
it into a network whose representations can be analyzed exactly. The two
approaches approximately agree with each other and with measurements on pretrained LLMs,
as illustrated in Fig.~\ref{fig:intro}(c).

Technically, several metrics have been introduced to quantify the
similarity of representations in LLMs. Recent work
\citep{groger2026revisiting} shows that global ones, including Centered
Kernel Alignment (CKA), based on cross-correlations between
representations \citep{kornblith2019similarity}, in fact provide limited
support for the Platonic hypothesis. By contrast, local methods, which
measure how well distances in one representation predict the neighborhood
structure of another, do support it and thus are favored here. These include mutual $k$-Nearest
Neighbors (mKNN) \citep{huh2024platonic} -- which, however, suffers from an
undesirable dependence on sample size \citep{koepke2026back} -- and the
Information Imbalance (II) \citep{glielmo2022ranking}. The
II measures how highly the nearest neighbors in one representation are
ranked in the distance ordering of another: it vanishes for perfectly
similar representations, and reaches a baseline of unity when
representations are unrelated. Detailed definitions of these metrics are
provided in Appendix~\ref{secapp:similarity_measures}.
II and mKNN lead to consistent results.

\vspace{-0.4em}
\section{Related work}

\paragraph{Transformers and synthetic grammars.} Transformers trained on
CFG-generated data have been argued to parse:
\citet{zhao2023transformers} show by construction that BERT-like models can approximate the inside-outside algorithm---which comprises an upward (encoding) and a downward (decoding) phase---and identify parsing information in trained models. Related conclusions hold for next-token prediction \citep{allen-zhu2025physics}. \citet{garnier-brun2025how} additionally study how hierarchical inference emerges during training, showing that correlations at successively larger scales are learned sequentially. For frozen tree topologies, \citet{zhou2026how} showed that the encoding
sweep suffices (the decoding phase collapsing into a single final layer) and that trained transformers can find this solution.
These works characterize the learned algorithm but not how it is learned,
and none addresses the multilingual setting.

\paragraph{The Random Hierarchy Model (RHM).} Random ensembles of grammars were
introduced by \citet{degiuli2019random}; frozen-tree hierarchical generative
models were proposed to study the learnability of deep functions
\citep{mossel2016deep, malach2018provably}. The RHM \citep{cagnetta2024how}
combines both -- a frozen tree with random production rules -- making learning
itself tractable: it predicts sample complexities, that production rules are
learned bottom-up \citep{cagnetta2024towards}, and neural scaling laws
tested on natural language \citep{cagnetta2026scaling}; it extends to
varying topologies \citep{parley2026deep}. For diffusion models learning  RHM data,
belief propagation is the optimal denoiser, predicting a phase transition
at finite noise observed in images and text
\citep{sclocchi2025phase, sclocchi2025probing}. Yet the multilingual
setting had not been considered,  facts 1--5 are unexplained.

\vspace{-0.4em}
\section{ Multilingual/Multimodal Random Hierarchy Model (MRHM)}
\label{sec:multilingual_rhm}


\paragraph{Generative grammars and the Random Hierarchy Model.}
A CFG generates an observed sequence by recursively expanding latent symbols according to production rules, making a generative tree whose leaves constitute the observed tokens \citep{chomsky1956three}. A probabilistic CFG additionally specifies a distribution over the possible production rules and therefore over sentences and their latent derivations.
The RHM is a controlled, finite-depth instance of this framework \citep{cagnetta2024how}. It generates an $s$-ary tree of depth $L$. Each node carries a symbol from an alphabet $\cA$ of size $v$; level $L$ contains the root, while level $0$ contains the observed tokens. 
We denote the level-$0$ leaf at position $i$ by
$x_i=X_i^{(0)}$, for $i=1,\ldots,s^L$, and denote its
level-$\ell$ ancestor by $X_i^{(\ell)}$. Thus,
$X_i^{(\ell)}=X_j^{(\ell)}$ whenever $x_i$ and $x_j$ belong to the same
level-$\ell$ sub-tree. A sketch of the RHM tree with notations is available in Appendix~\ref{secapp:rhm_schematic}.

At each level, a production rule maps a parent symbol $a \in \cA$ to an ordered tuple of children $(b_1,\ldots,b_s)\in\cA^s$. Each parent has $m$ possible rules, giving $mv$ rule tuples per level. The tuples are sampled without replacement from $\cA^s$, ensuring that each child tuple maps back to a unique parent. A sentence is generated by sampling a root symbol and recursively selecting one of the $m$ rules available to each node. The RHM is therefore specified by $(L, s, m, v)$; in Fig.~\ref{fig:intro}(a) $L=3$ and $s=2$.

\paragraph{Multilingual RHM.}
In natural language, translations share the abstract content while expressing it through different language-specific structures. This is also true for paired images and captions. We model these two situations  by considering two RHM grammars that generate two languages $A$ and $B$. The languages share the upper $L_{\mathrm{shared}}$ levels of the generative hierarchy (same production rules and symbols) but use distinct production rules and distinct symbols in the lower $L_{\mathrm{surface}}$ levels, with $L=L_{\mathrm{shared}}+L_{\mathrm{surface}}$. Thus at the split level $L_{\mathrm{surface}}$ (and only there), each symbol has $2m$ outgoing rules, with $m$ rules for each language, giving $2mv$ rules in total.

A translation pair is generated by sampling a common upper tree down to the split level $L_{\mathrm{surface}}$ and then expanding the lower subtrees using the rules of languages $A$ and $B$, respectively. The resulting sentences therefore share all the latent variables at levels $\ell\geq L_{\mathrm{surface}}$ but differ below this split. We interpret the shared variables as common semantic content and the lower variables as its language-specific realization. Consequently, $L_{\mathrm{surface}}$ provides a controlled proxy for language distance: increasing $L_{\mathrm{surface}}$ reduces the amount of shared generative structure.

\paragraph{Code-switching as a cross-lingual training signal.}
Multilingual corpora contain mixed-language contexts that can provide models with evidence connecting otherwise distinct linguistic realizations. Consistent with the evidence that synthetic code-switching can improve cross-lingual alignment \citep{wang2025investigating}, we introduce a controlled form of code-switching during pretraining.
For each level-$L_{\mathrm{surface}}$ sub-tree in a language-$A$ sentence, we replace its language-$A$ realization with the corresponding language-$B$ realization with a probability $\epsilon$, independently for all sub-trees; and analogously from $B$ to $A$.


\section{Theory: Next-token Prediction  via Belief Propagation}
\label{sec:bp}

Because the RHM has a tree-structured dependency graph, the Bayes-optimal next-token distribution
$p(x_{t+1}\mid x_{1:t})$ can be computed by belief propagation (BP) \citep{sclocchi2025phase}. This remains true in the multilingual setting. As illustrated in Figure \ref{fig:intro}(b), the computation occurs in two stages: encoding (blue nodes) and decoding (orange nodes). 
During encoding, evidence from the observed tokens propagates upward through the hierarchy, producing beliefs over the latent variables. During decoding, beliefs propagate downward along the path to $x_{t+1}$, yielding its posterior predictive distribution.


To compare this computation with trained transformers, we construct theoretical networks whose layerwise representations linearly encode the BP messages. Belief over a latent symbol is represented by a vector in $\mathbb{R}^v$, with one coordinate for each symbol in the alphabet $\cA$. Each network layer implements one sequential step of the BP computation. 


\textbf{Encoding.}
At token position $t$, the message $\bu_t^{(\ell)}\in\mathbb{R}^v$ represents the belief over the latent variable $X_t^{(\ell)}$, inferred from the observed tokens within its subtree.
The unobserved tokens are marginalized. The messages are computed recursively, as described in Appendix~\ref{secapp:bp}.

For the language-specific levels $\ell<L_{\mathrm{surface}}$, beliefs for the two languages occupy separate $v$-dimensional orthogonal blocks, giving a $2v$-dimensional representation. Separate blocks also occur for the shared levels $\ell\geq L_{\mathrm{surface}}$; however, we will empirically show and quantify in \S~\ref{sec:coordinate_alignment} that some amount of coordinate alignment -- a finite scalar product -- can occur depending on the code-switching rate $\epsilon$ and the training set size $P$. As shown in Appendix~\ref{secapp:coordinate_alignment}, however, such alignment has little effect on geometric, neighborhood similarity; thus in the main text theoretical networks assume orthogonal representations for simplicity.  
The layerwise representations of the theoretical network accumulate the messages computed up to that point:

\begin{align}
\label{eq:bp_activations_encoder}
\bh_t^{(k)}
    &= \bigoplus_{\ell=1}^{k}\bu_t^{(\ell)},
    & 1 \leq k \leq L-1,
\end{align}
where $\bigoplus$ denotes concatenation. The messages are accumulated at different levels because they are subsequently used during decoding.\footnote{A more parsimonious version of BP would only encode the messages $u^{(l)}_t$ actually used by the decoder. 
We show in Appendix \ref{secapp:bp} that such distinctions do not qualitatively change our findings.} 

\textbf{Decoding.}
Once the observed prefix has been encoded, information propagates downward along the ancestral path to the token being predicted. At level $\ell$, the downward message $\bd^{(\ell)}_t$ represents the posterior distribution of $X^{(\ell)}_{t+1}$, conditioned on the entire observed prefix $x_{1:t}$. Its recursive computation is given in Appendix~\ref{secapp:bp}. As in the encoding stage, the beliefs below $L_{\mathrm{surface}}$ are language-specific, whereas those at and above $L_{\mathrm{surface}}$ are shared.



After the computation of a downward message, the corresponding upward message is no longer needed, and can be discarded. However, in an LLM there is no cost in keeping useless messages, as long as they do not interfere with the downstream computations.
To allow for this possibility, we gradually decay used messages with a multiplicative factor $\gamma \in [0,1]$ at each step after usage. The resulting layerwise representations are:
\begin{align}
\label{eq:bp_activations_decoder_gamma}
\bh_t^{(k)}
    &= \bd_t^{(2L-1-k)} \oplus \underbracket{\left( \bigoplus_{\ell=1}^{2L-2-k} \bu_t^{(\ell)} \right)}_{\mathrm{not\, yet\, used}} \oplus \underbracket{\left(\bigoplus_{\substack{\ell=2L-1-k \\ \ell\neq 0}}^{L-1} \gamma^{\ell-(2L-2-k)} \bu_t^{(\ell)} \right)}_{\mathrm{used}}
       ,
    & L \leq k \leq 2L-1.
\end{align}
\vspace{-0.7em}

Instantaneous removal of used messages corresponds to $\gamma=0$. As shown in Fig.~\ref{figapp:layerwise_prediction} of Appendix~\ref{secapp:bp},  although the value of $\gamma$ does not alter qualitative findings, $\gamma=1/2$ leads to representations that match those found in language models trained on the RHM significantly better. 
Overall, the final message $\bd_t^{(0)}$ is precisely the predictive distribution $p(x_{t+1}\mid x_{1:t})$. The complete theoretical network contains $2L-1$ layers, one for each step of the encoder--decoder computation.

\textbf{Incomplete inference.}
Transformers approach optimal performance on RHM data only when trained on
sufficiently many samples. With a smaller training set, they recover the
production rules only up to an effective depth $\ell_\ast<L$, which increases
with the number of training examples $P$ \citep{cagnetta2024towards}. 
This produces suboptimal next-token predictions. 
We model this regime using an incomplete BP algorithm that only recovers
production rules up to level $\ell_\ast$, and higher level latents are not reconstructed. Note that a transformer has to perform fewer computational steps in this case, and passive layers will appear. For simplicity, we assume that they are the middle layers. Our theoretical results depend on this choice quantitatively, but not qualitatively. Details of this construction are discussed in Appendix~\ref{secapp:bp}.

\textbf{Alternative algorithm.}
Fixed-topology grammars also admit an $(L{+}1)$-step shortcut in which the downward computation collapses into a single final update \citep{zhou2026how}, while the encoder is exactly the same. We refer to this algorithm and to the associated theoretical network as ``encoder-only". Depending on the training set size ($P$) and the code-switching rate ($\epsilon$), trained transformers exhibit representations consistent with either the encoding-decoding or encoder-only algorithm. We expect encoder-decoder to be the robust solution that characterizes natural languages\footnote{Grammars of natural languages have varying tree topology. In Appendix~\ref{secapp:varying_topology}, we show empirically that transformers trained on such grammars  indeed display an encoder-decoder architecture \citep{parley2026deep}.}.
We therefore focus on $\epsilon=0.1$, for which trained transformers consistently exhibit this structure, and study the transition between the two algorithms in Appendix~\ref{secapp:transition}.



\section{Geometric similarity}

\subsection{Transformer training setup and similarity measure}
We train GPT-style decoder-only transformers for next-token-prediction on MRHM data.
Unless stated otherwise, the data are sampled from a MRHM with $L=4, s=2, m=4, v=16$, and two languages with split levels $(L_{\mathrm{shared}}, L_{\mathrm{surface}}) \in \{(3,1), (2,2), (1,3)\}$. The transformers have 8 layers, embedding dimension 512, and 4 attention heads.\footnote{For $L=4$, BP has $2L-1=7$ sequential steps. Appendix~\ref{secapp:varying_n_layers} shows that our conclusions are
unchanged for any choice of $n_{\mathrm{layers}}\geq2L-1$; additional layers simply stretch
the layerwise profile across network depth.}
We trained on datasets containing $P$ sentences using cross-entropy loss and the AdamW optimizer for multiple epochs until convergence. For each run, we select the checkpoint with the lowest validation loss and perform all analyses on a held-out dataset not used during training. We perform similarity measurements on token-averaged representations of the networks.


\subsection{Geometric similarity in multilingual models}


We compare the geometric similarity in transformers trained on multiple languages with that of the theoretical networks constructed from BP (\S \ref{sec:bp}). Both networks are evaluated on the same collection of translation pairs. We study this for both: a single network trained on two languages and independent networks separately trained on the two languages respectively.


\begin{figure}[!h]
    \centering
    \includegraphics[width=0.99\linewidth]{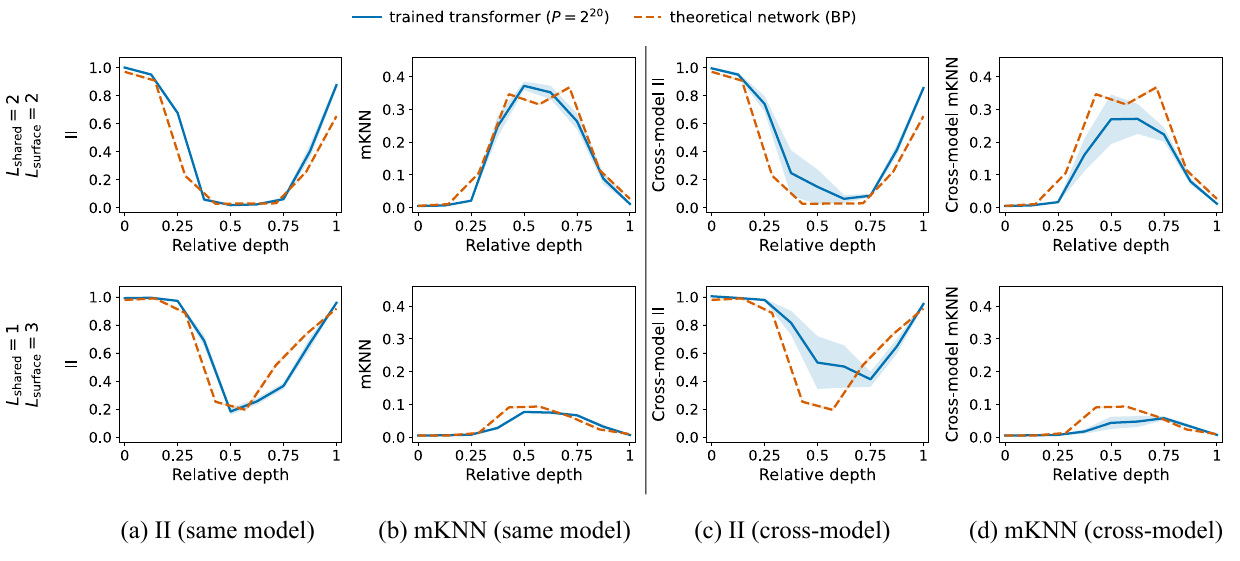}
    \vspace{-0.3em}
    \caption{\textbf{(a,b)} Representations of a transformer trained on MRHM are geometrically similar in the middle layers when processing pairs of translated sentences. The geometric similarity is stronger for structurally similar languages, controlled by $L_{\mathrm{surface}}$ in our setup. \textbf{(c,d)} This similarity also exists between representations of independent models, trained separately on languages A and B. Measurements with II
    and mKNN show similar trends. In all cases, the empirical trends are well-captured by theoretical networks constructed with BP. 
    }
    \label{fig:similarity}
    \vspace{-1em}
\end{figure}

\paragraph{Geometric similarity is not perfect and peaks in the middle layers (1 \& 2).}

Across different MRHM settings ($L_{\mathrm{surface}}$), the representations of translation pairs exhibit a non-monotonic similarity profile along transformer layers (Figure \ref{fig:similarity}). Similarity is low near the input, increases substantially in the middle layers (but does not become perfect), and decreases again near the output; producing a U-shaped profile in II, and an inverted-U shaped profile in mKNN -- as shown in Fig.~\ref{fig:similarity} (a) and (b), respectively. 

The corresponding BP networks reproduce the profiles qualitatively. This layerwise organization follows directly from the structure of inference, as sketched in Fig.~\ref{fig:intro}~(b). During the initial encoding steps, the BP messages depend on the language-specific lower levels of the grammar and similarity of translations is absent. As implied by Eqs.~\ref{eq:bp_activations_encoder} and~\ref{eq:bp_activations_decoder_gamma}, the middle layers encode the high-level latents that are shared by translated pairs, leading to the observed similarity. The latter is not perfect, since the information on low-level latents is still needed to ultimately be used in the decoder to predict the next token. 
Finally, during the final decoding steps, the computation returns to language-specific rules to predict the next token, causing the similarity to decrease in the final layers.

\textbf{Geometric similarity strengthens with language similarity (3).}
The magnitude of middle-layer similarity depends systematically on the amount of generative structure shared between languages. Increasing the language-specific depth $L_{\mathrm{surface}}$, while correspondingly decreasing the shared depth $L_{\mathrm{shared}}$, raises the minimum II and lowers maximum mKNN (see Fig.~\ref{fig:similarity}). This is expected from the theory, since in the middle layers a larger part of the representation is shared across translated pairs.

\textbf{Separately trained models also exhibit geometric similarity.} 
Training two independently initialized transformers, one on language-A sentences and the other on language-B, exhibits similar layerwise similarity trends as a single trained model (Fig.\ref{fig:similarity}(c,d)).


\vspace{-0.5em}
\begin{figure}[!h]
    \centering
    \includegraphics[width=\linewidth]{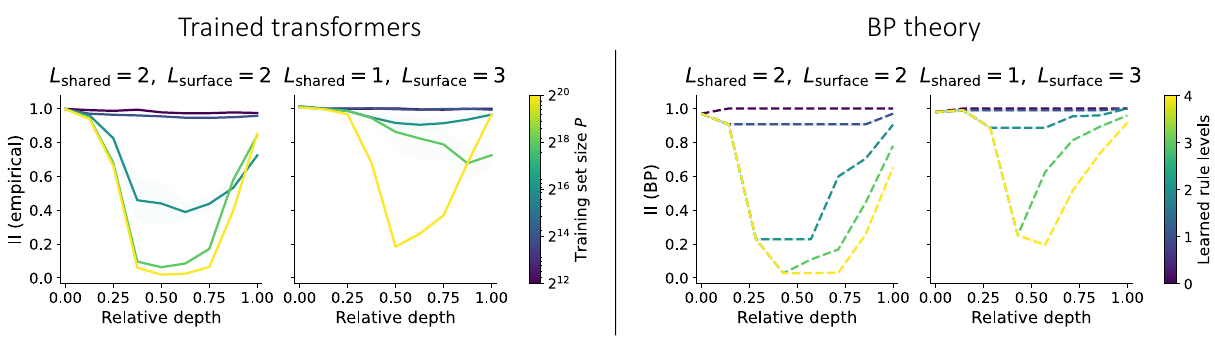}
    \caption{\textbf{(Left)} Cross-lingual similarity gets stronger as the training-set size $P$ increases. \textbf{(Right)} Our theory captures this behaviour with the imperfect-BP algorithm, which only knows the RHM rules up to level $\ell_\ast$. 
    }
    \label{fig:ii_with_P}
\end{figure}

\textbf{Geometric similarity correlates with performance (4).}
At small training-set sizes, transformers exhibit poor next-token-prediction performance and little cross-lingual similarity. As $P$ increases, geometric similarity progressively emerges in the middle layers, as indicated by lower II (Fig.~\ref{fig:ii_with_P}).

This behaviour is explained by the imperfect-BP construction of \S \ref{sec:bp}. With finite data, the network learns the production rules only up to level $\ell_\ast$. When $\ell_\ast$ does not extend into the shared portion of the hierarchy, the model cannot construct geometrically similar high-level representations. Middle-layer similarity emerges once the learned computation reaches the first shared level, and becomes stronger as additional shared levels are recovered. The discrete progression with $\ell_\ast$ in the theoretical networks reproduces the continuous improvement with $P$ observed in the trained transformers.

\begin{figure}[!h]
    \centering
    \includegraphics[width=\linewidth]{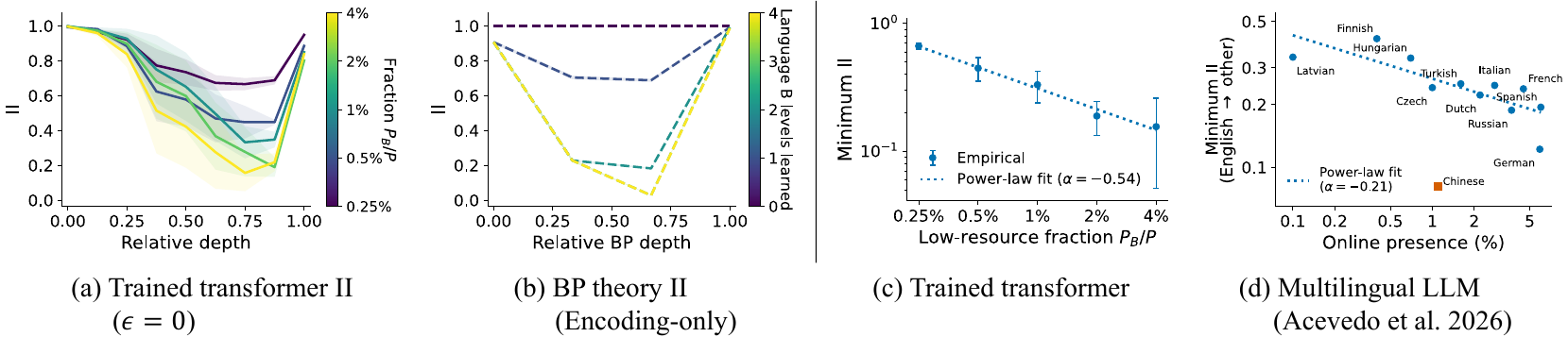}
    \caption{Cross-lingual geometric similarity with low-resource languages. 
    \textbf{(a)} Similarity as measured by II gets weaker as the frequency of language-B decreases ($\epsilon=0$).
    \textbf{(b)} BP theory captures this behaviour by considering fully learned language-A and partially learned language-B, up to level $\ell$.
    \textbf{(c)} The minimum II (across layers) decreases monotonically with the fraction of low-resource language ($P_B/P$).
    \textbf{(d)} Plot adapted from \citep{acevedo2026quantitative}: Multilingual LLM shows qualitatively similar trend -- minimum II between English and other languages decreases monotonically with the online presence of the other language.
    Both theory and observations display apparent power law behavior in the range probed with exponents $\alpha\approx 0.5$ and $\alpha\approx 0.2$ respectively.
    \vspace{-0.5em}
    }
    \label{fig:low_resource}
\end{figure}

\textbf{Geometric similarity weakens for low-resource languages.}
To capture the effect of language-specific exposure in pretraining, we fix the total dataset size ($P$) and vary the fraction of samples from each language, $P_A/P$ and $P_B/P$, with $P_A+P_B=P$. 
To precisely control the sample fractions, we set $\epsilon=0$ for this experiment. Fig.~\ref{fig:low_resource}(a) shows that geometric similarity gets stronger as the low-resource fraction $P_B/P$ is increased. BP theory reproduces this trend, with stepwise improvements as more language-B data allows additional grammar levels to be resolved (Fig.~\ref{fig:low_resource}(b)). Note that here we use the $(L{+}1)$-step encoder-only BP computation, to match the $\epsilon=0$ case. 
Fig.~\ref{fig:low_resource}(c) shows the maximal geometric similarity (minimum II) across layers, as a function of $P_B/P$. A clear monotonic decrease of the minimum II shows that the similarity between two languages depends significantly on how much data from each language the model sees during training.
These results are consistent with experiments  performed on natural language in \citet{acevedo2026quantitative}, where the abundance of languages in training corpora was approximated with their online presence. We show their results in Fig.~\ref{fig:low_resource}(d) for comparison.

\vspace{-0.5em}
\section{Layerwise Novelty}


\textbf{Definition of novelty.} Can simple manipulations of LLM representations give more direct access to shared semantic content? Our theory implies that they can. As indicated in Eq.\ref{eq:bp_activations_encoder}, at each step of the encoding phase the representation contains a newly computed higher-level message, together with earlier lower-level language-specific messages, retained for subsequent decoding. Because the retained messages are predictable from the preceding layer, subtracting this predictable component
isolates the more abstract information introduced by the current layer. We call this residual \emph{layerwise novelty} and predict that it exposes shared computation more clearly than the raw hidden states. Details of the procedure to compute layerwise novelty are in Appendix~\ref{secapp:novelty}.

\begin{figure}[!h]
    \centering
    \includegraphics[width=0.95\linewidth]{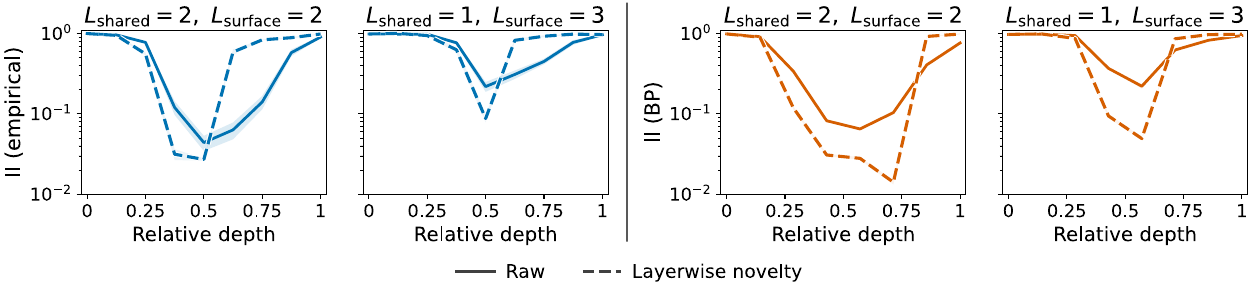}
    \caption{In both trained transformers and theoretical BP networks,
    layerwise novelty reveals more strongly similar intermediate
    computations than raw hidden states. 
    }
    \label{fig:novelty}
    \vspace{-1em}
\end{figure}

\textbf{Novelty improves cross-lingual similarity in the RHM.}
We first test our prediction in the MRHM. Fig.~\ref{fig:novelty} confirms that layerwise novelty is more similar across languages at intermediate layers than raw representations. Novelty lowers the II in the first part of the network, including the minimum II at the middle layer. In later layers, novelty shows higher II because it removes the high-level information carried by the skip connections.
Similarity measured with mKNN shows comparable results (Fig.\ref{figapp:rhm_novelty}). 
These results indicate that shared structure is concentrated in the information computed in the intermediate layers, but is partially obscured in raw hidden states by language-specific information carried forward from earlier layers.

\begin{figure}[ht]
    \centering
    \includegraphics[width=0.95\linewidth]{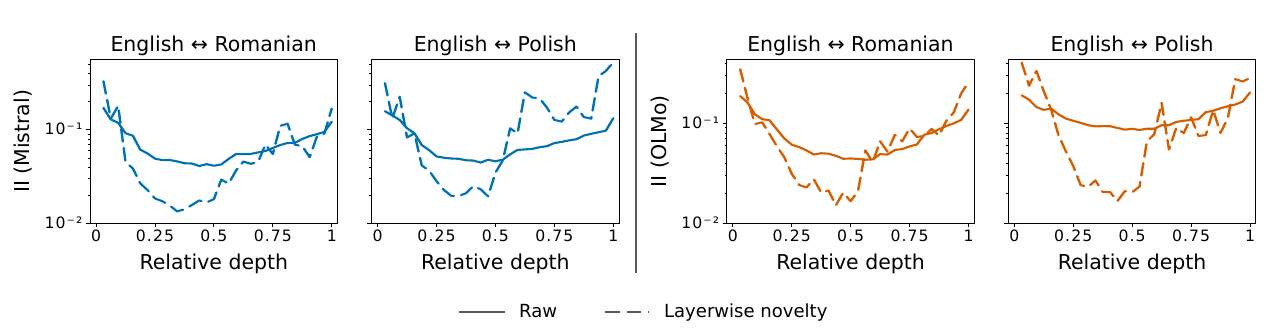}
    \caption{
        Comparison of II on raw representations and layer-wise novelty  in Mistral-7B-v0.1 and OLMo-3-7B for parallel English--Romanian and English--Polish sentence pairs from Europarl. } 
    \vspace{-1em}
    \label{fig:novelty_natural_language}
\end{figure}

\textbf{Layerwise novelty improves cross-lingual similarity in natural languages.}
We test this prediction in pre-trained language models, taking representations from Mistral-7B-v0.1 \citep{jiang2023mistral7b} and OLMo-3-7B on parallel English--Romanian and English--Polish sentence pairs from Europarl \citep{koehn2005europarl}. In Fig.~\ref{fig:novelty_natural_language}, we show that layerwise novelty achieves lower II than raw representations across most layers. Furthermore, we confirm in Appendix~\ref{secapp:novelty} that these results (i) hold for other models/language-pairs, (ii) hold for the mKNN metric, (iii) outperform the simple removal of residual connections, and (iv) outperform the supra-word residual \citep{toneva2022combining} -- an alternative method obtained by subtracting from contextual embeddings the component linearly predictable from individual word embeddings.

\vspace{-0.5em}
\section{Coordinate Alignment }
\label{sec:coordinate_alignment}
\vspace{-0.5em}

Geometric similarity does not imply coordinate alignment. Geometric similarity is a statement
about shape: it compares distances and neighborhoods, and is therefore
meaningful even when the two representations live in different spaces -- it can be measured across separate models. To observe coordinate alignment, shared information must be encoded along the same directions of a common space, so that a readout learned on one language operates on the other without refitting. It is coordinate alignment, not geometric similarity, that underlies cross-lingual transfer. In this section we establish that the two properties are decoupled in our model: the shared hierarchy alone produces geometric similarity, whereas coordinate alignment requires an explicit cross-lingual signal -- it is controlled jointly by the training-set size $P$ and the code-switching rate $\epsilon$.

\vspace{-0.6em}
\begin{figure}[!h]
    \centering
    \includegraphics[width=0.7\linewidth]{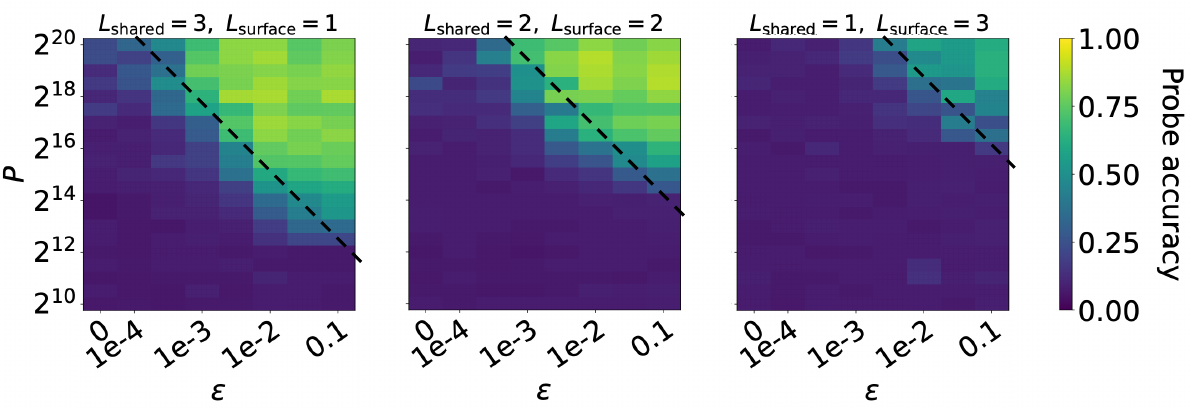}
    \vspace{-0.5em}
    \caption{Alignment measured by latent cross-probing. Linear probes for
    latents at level-$L_{\mathrm{surface}}$ are trained on language-A
    representations and tested on language-B representations. Accuracy is
    controlled jointly by the code-switching rate $\epsilon$ and the
    training-set size $P$: the aligned region (green) appears along the
    predicted boundary $P_{\mathrm{threshold}}\propto\epsilon^{-1}$ (dashed line).}
    \label{fig:latent_probing_epsilon}
    \vspace{-0.6em}
\end{figure}

We measure alignment using latent cross-probing: a linear probe is trained
on language-A representations to predict a latent variable of the shared
hierarchy, then evaluated, without refitting, on language-B
representations. High cross-probing accuracy indicates that the two
languages encode the shared latent in compatible coordinates. As shown in
Figure~\ref{fig:latent_probing_epsilon}, cross-probing accuracy depends
jointly on $\epsilon$ and $P$. At fixed $P$, increasing $\epsilon$
strengthens alignment, thus explaining (5). At fixed $\epsilon>0$, increasing $P$ allows the
model to exploit the increasingly well-sampled cross-lingual signal. In
contrast, at $\epsilon=0$, increasing the amount of data alone
does not produce substantial cross-probing accuracy: the model becomes
geometrically similar across languages, yet never aligned.

This sample complexity has a simple interpretation. In monolingual
RHM, building aligned representations of latents requires seeing pairs of features in the training set a sufficient number of times  \citep{cagnetta2024towards}. Here, features from different languages  are observed only through code-switched subtrees, and are
therefore suppressed by a factor proportional to $\epsilon$: a dataset of
size $P$ carries an effective cross-lingual signal of order $\epsilon P$.
Reaching a fixed level of alignment requires $\epsilon P$ to remain
constant, predicting $P_{\mathrm{threshold}} \propto \epsilon^{-1}$,
in agreement with the boundary of the aligned region in
Fig.~\ref{fig:latent_probing_epsilon} (dashed line).

\vspace{-0.6em}
\section{Conclusion}
\vspace{-0.5em}
We have shown that a single assumption --that data possess a hidden hierarchy,
with semantic content residing at its top and language-specific structure at
its bottom-- accounts for the phenomenology of multilingual representations:
similarity peaking in the middle layers, coexisting with language-specific
structure, strengthening with language proximity, model quality and data
exposure, and distinct from the stronger property of alignment, which
requires cross-lingual signal in the training data. Since the theory
is not specific to language, two modalities can be viewed as two surface
realizations of a shared upper hierarchy, suggesting that the same analysis
applies to the cross-modal similarities motivating the platonic
representation hypothesis. It also carries a practical implication:
alignment is purchased with mixed-language signal, identifying code-switched data as the efficient lever for
low-resource languages.

A central consequence of the theory is constructive: it prescribes how to
build more abstract representations, by removing from each layer what is
predictable from the previous one -- eliminating surface details to isolate
newly computed, higher-level information. This layerwise novelty may prove
useful in  AI to study how abstractions are processed, but also in neuroscience. Indeed,  LLM representations are currently the best predictors of brain activity during language processing \citep{caucheteux2022brains}, 
and novelty may offer a principled,
depth-resolved improvement.

\subsubsection*{Acknowledgments}

We thank Marco Baroni and Alessandro Laio for insightful discussions throughout the project and feedback on the manuscript; Jack Parley for discussions on varying topology and for sharing the code used in those experiments; Francesco Cagnetta, Sophia Houhamdi, Jean-R\'emi King,
Alessandro Pacco, and the members of the Simons Collaboration on the Physics of Learning and Neural Computation for useful discussions; and Francesco D'Amico, Alessandro Favero, and Laurent Bugnard for feedback on the manuscript.
This work was supported by the Simons Foundation through the Simons Collaboration on the Physics of Learning and Neural Computation (Award ID: SFI-MPS-POL00012574-05), PI Wyart. Computational resources were provided by the Skipjack cluster, operated by Advanced Research Computing at Hopkins (ARCH). Darshil Doshi acknowledges financial support from the Johns Hopkins University Data Science and AI Institute Fellowship. Daniel Korchinski acknowledges financial support from the Natural Sciences and Engineering Research Council of Canada (NSERC PDF - 587940 - 2024). Santiago Acevedo acknowledges financial support by the region Friuli Venezia Giulia (project F53C22001770002).

\bibliography{iclr2027_conference}
\bibliographystyle{iclr2027_conference}

\newpage 
\appendix

\section{RHM Schematic}
\label{secapp:rhm_schematic}

\begin{figure}[!h]
    \centering
    \includegraphics[width=0.9\linewidth]{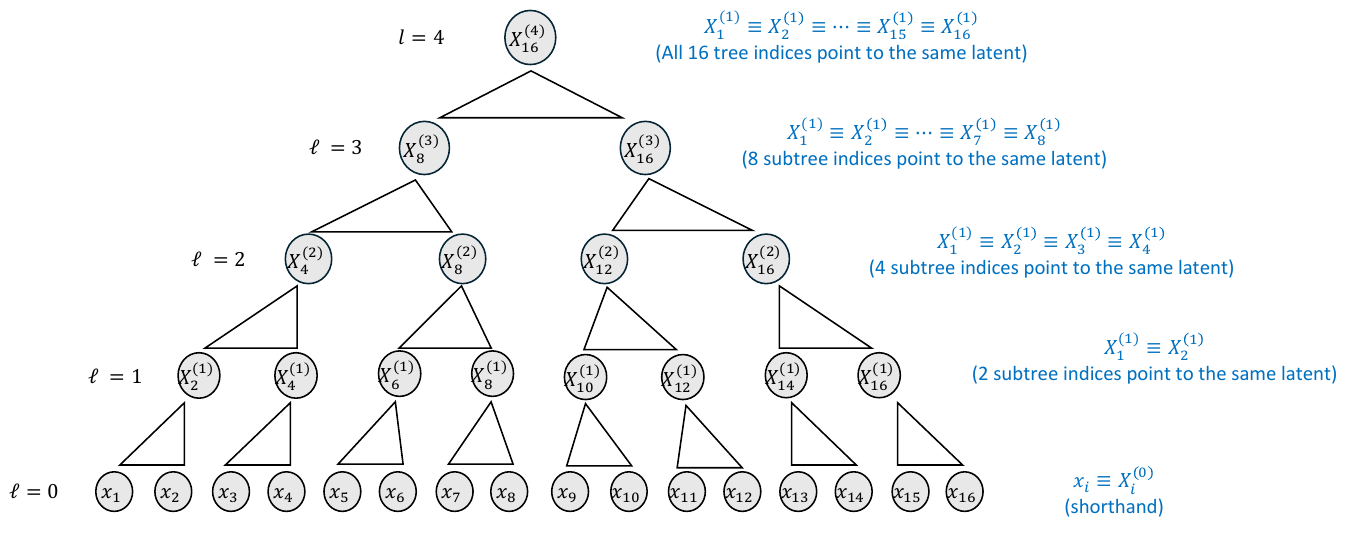}
    \caption{Schematic of RHM with $L=4$. Visible tokens, i.e. leaves, are denoted as $x_i \equiv X^{(0)}_i$. Level-$\ell$ parent node of $i$-th token is denoted as $X^{(\ell)}_i$ -- this introduces a redundancy in notation since this latent can be denoted with any of the $2^\ell$ subscripts, corresponding to the tokens in its subtree.}
    \label{figapp:rhm_schematic}
\end{figure}

\section{Belief Propagation and Theoretical Networks}
\label{secapp:bp}

\begin{figure}[!h]
    \centering
    \includegraphics[width=\linewidth]{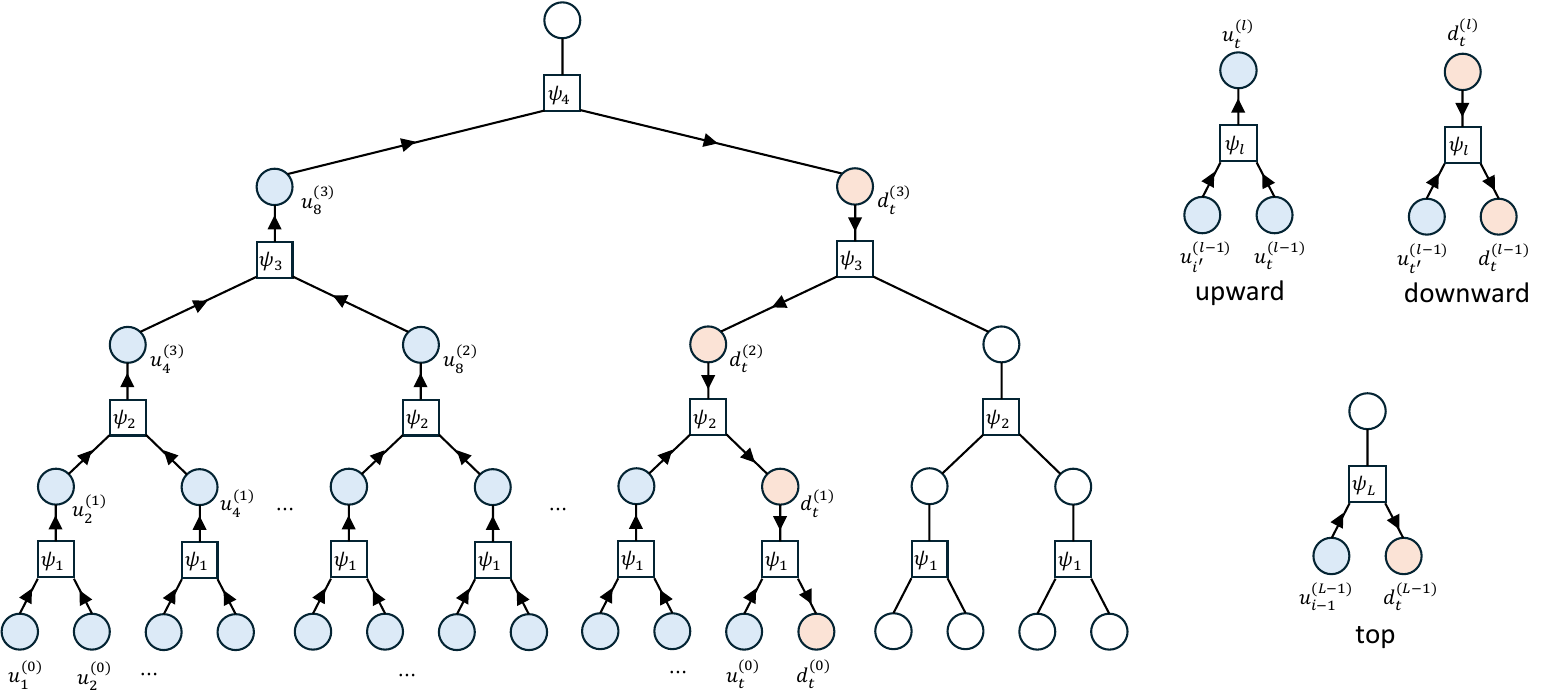}
    \caption{\textbf{Belief propagation for causal next-token prediction.}
    Blue nodes denote upward messages summarizing evidence within observed
    subtrees; orange nodes denote downward messages along the path to
    $x_{t+1}$. Unfilled nodes are unobserved, and squares denote grammar
    factors $\psi_\ell$. The right panels show the local upward, downward,
    and top-level updates.}
    \label{figapp:bp_schematic}
\end{figure}

We describe the belief-propagation (BP) algorithm underlying
\S~\ref{sec:bp} and its representation as a theoretical network.
Figure~\ref{figapp:bp_schematic} illustrates the computation: upward messages
summarize evidence within observed subtrees, while downward messages propagate
information along the boundary leading to $x_{t+1}$.

\paragraph{Notation and grammar factors.}
Each message is a vector in $\mathbb{R}^v$. Its level and position,
$(\ell,i)$, uniquely identify the corresponding latent node
$X_i^{(\ell)}$. When specifying the BP recursions, we write
$[\bu_i^{(\ell)}]_a$ for the component associated with the assignment
$X_i^{(\ell)}=a$. Thus, $a$ indexes a component of the message. We use $i$ for a generic position; setting $i=t$
recovers the causal-boundary messages used in the main text.

For $1\leq\ell\leq L-1$, let $\psi_\ell(b,\ba)$ indicate whether
$b\rightarrow\ba$ is a valid production rule, where $b\in\cA$ and
$\ba=(a_1,\ldots,a_s)\in\cA^s$:
\begin{equation}
\label{eq:bp_rule_factor}
\psi_\ell(b,\ba)
=
\begin{cases}
1, & b\rightarrow\ba
     \text{ is a valid rule at level }\ell,\\
0, & \text{otherwise}.
\end{cases}
\end{equation}
Uniform rule probabilities can be omitted because all messages are
normalized.

The top factor is root-agnostic:
\begin{equation}
\label{eq:bp_top_factor}
\psi_L(\ba)
=
\begin{cases}
1, & \ba \text{ is a valid top-level tuple},\\
0, & \text{otherwise}.
\end{cases}
\end{equation}
Indeed, sampling one of $v$ root symbols followed by one of its $m$ rules is
equivalent to using a single dummy root with $vm$ possible expansions. Thus,
no prior or message over the root is required.

The updates below $L_{\mathrm{surface}}$ use language-specific production
rules. At the split, a shared parent has $m$ admissible expansions in each
language, while the production rules above the split are shared. All messages
below depend implicitly on the observed prefix $x_{1:t}$.

\paragraph{Upward messages.}
Consider a level-$\ell$ node $X_i^{(\ell)}$ with ordered children
$X_{i_1}^{(\ell-1)},\ldots,X_{i_s}^{(\ell-1)}$. Its upward message summarizes
the evidence within its subtree:
\begin{equation}
\label{eq:bp_upward_recursion}
\left[\bu_i^{(\ell)}\right]_b
=
\sum_{\ba\in\cA^s}
\psi_\ell(b,\ba)
\prod_{j=1}^s
\left[\bu_{i_j}^{(\ell-1)}\right]_{a_j},
\qquad
1\leq\ell\leq L-1.
\end{equation}
The recurrence is initialized at the leaves by
\begin{equation}
\label{eq:bp_leaf_initialization}
\left[\bu_i^{(0)}\right]_a
=
\begin{cases}
\delta_{a,x_i}, & i\leq t,\\
1/v,            & i>t.
\end{cases}
\end{equation}
Observed leaves therefore provide point-mass evidence, while unobserved leaves
are marginalized. 

At transformer position $t$, $\bu_t^{(\ell)}$ denotes the message for the
level-$\ell$ subtree intersected by the causal boundary. If this subtree is
only partially observed, Equation~\ref{eq:bp_leaf_initialization} marginalizes
its unobserved leaves. Strictly, $\bu_t^{(\ell)}$ is the likelihood of the
observed descendants as a function of the latent symbol; we refer to its
normalized form as a belief.

\paragraph{Downward messages.}
The downward computation follows the path from the top of the hierarchy to
$x_{t+1}$. At each factor, let $a$ denote the symbol of the child on this
path and let $\ba_{\mathrm{sib}}=(a_1,\ldots,a_{s-1})$ denote its siblings.
The position of the target child within the ordered tuple is fixed by $t$ and
is left implicit. We write $\bu_{\mathrm{sib},j}^{(\ell)}$ for the upward
message from its $j$-th sibling.

The first downward update is performed directly by the root-agnostic factor:
\begin{equation}
\label{eq:bp_downward_top}
\left[\bd_t^{(L-1)}\right]_a
=
\sum_{\ba_{\mathrm{sib}}\in\cA^{s-1}}
\psi_L(a,\ba_{\mathrm{sib}})
\prod_{j=1}^{s-1}
\left[\bu_{\mathrm{sib},j}^{(L-1)}\right]_{a_j}.
\end{equation}
Here $\psi_L(a,\ba)$ denotes the top-level factor evaluated
on the ordered tuple formed by the target child and its siblings. No root
prior or root message is required.

At lower levels, the update additionally incorporates the downward message
from the parent:
\begin{equation}
\label{eq:bp_downward_recursion}
\left[\bd_t^{(\ell-1)}\right]_a
=
\sum_{b\in\cA}
\sum_{\ba_{\mathrm{sib}}\in\cA^{s-1}}
\left[\bd_t^{(\ell)}\right]_b
\psi_\ell(b;\ba)
\prod_{j=1}^{s-1}
\left[\bu_{\mathrm{sib},j}^{(\ell-1)}\right]_{a_j},
\qquad
1\leq\ell\leq L-1.
\end{equation}

The downward message is formally a cavity belief containing evidence from
outside the target subtree. The leaf-level message is the next token posterior
\begin{equation}
\label{eq:bp_predictive_distribution}
\left[\bd_t^{(0)}\right]_a
=
p(x_{t+1}=a\mid x_{1:t}).
\end{equation}

\paragraph{Theoretical-network representations.}
The BP messages are assembled into layerwise representations as defined in
Equations~\ref{eq:bp_activations_encoder} and
\ref{eq:bp_activations_decoder_gamma}. We realize these direct sums in a
fixed ambient space by assigning each message level its own block and setting
inactive blocks to zero.

To avoid imposing coordinate alignment, we embed the two languages into
orthogonal subspaces. For a message vector $\mathbf m\in\mathbb{R}^v$, define
\begin{equation}
\label{eq:bp_orthogonal_embedding}
\iota_A(\mathbf m)
=
\begin{bmatrix}
\mathbf m\\
\mathbf 0
\end{bmatrix},
\qquad
\iota_B(\mathbf m)
=
\begin{bmatrix}
\mathbf 0\\
\mathbf m
\end{bmatrix}
\in\mathbb{R}^{2v}.
\end{equation}
We apply this convention even when $\mathbf m$ describes a shared latent
variable. It preserves the local geometry of the messages within each language
without placing the two languages in the same coordinates.

\begin{figure}[!h]
    \centering
    \includegraphics[width=\linewidth]{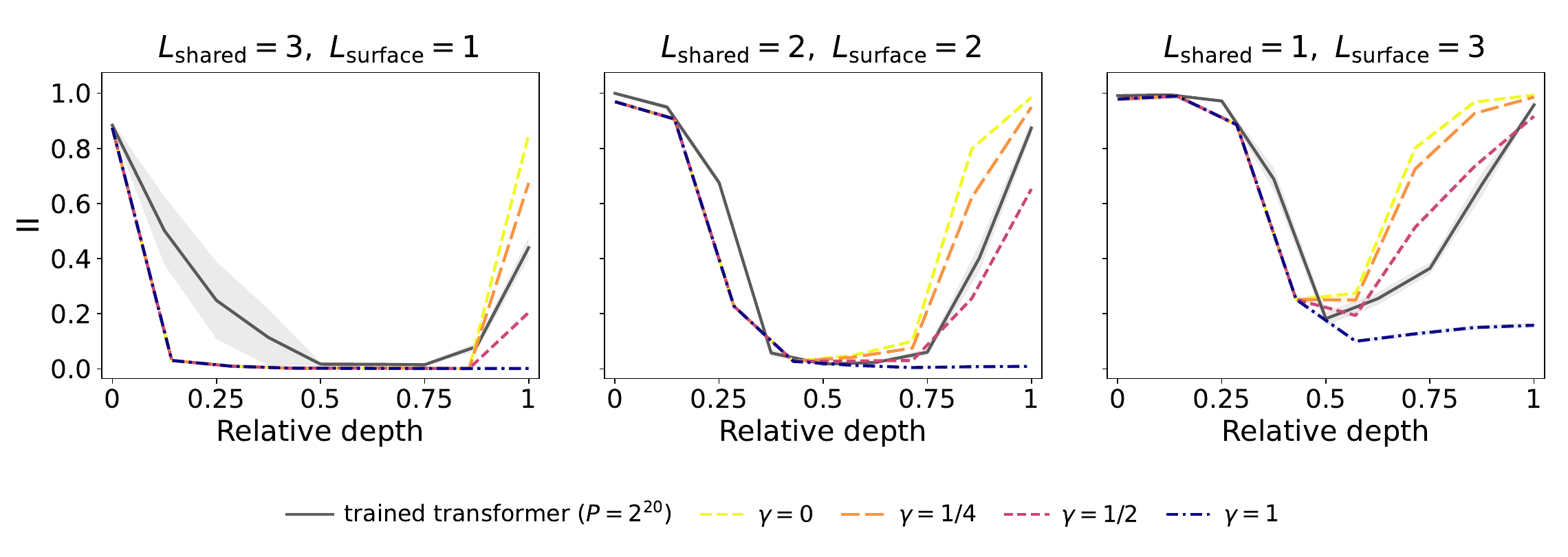}
    \caption{Comparing layerwise II profiles for different $\gamma \in [0,1]$. Empirical II curves from trained transformers shown for reference.}
    \label{figapp:II_with_gamma}
\end{figure}

\paragraph{Alternative message representations and algorithms.}
As discussed in the main text and Eq.~\ref{eq:bp_activations_decoder_gamma}, while constructing the later layers of the theoretical network we decay the upward messages after their usage with a factor $\gamma \in [0,1]$ at each step.
Here $\gamma=0$ recovers immediate deletion, whereas $\gamma=1$ retains all
upward messages. 
While this choice does not affect the qualitative results of the paper, the best quantitative match to trained transformers is achieved at intermediate values of $\gamma$. We make one such choice in the main text, namely $\gamma=1/2$. This is a direct consequence of the residual structure of transformers, where skip connections prevent immediate removal of messages. We compare the layerwise similarity behaviours for different $\gamma$ in Fig.~\ref{figapp:II_with_gamma}

\begin{figure}[!h]
    \centering
    \includegraphics[width=\linewidth]{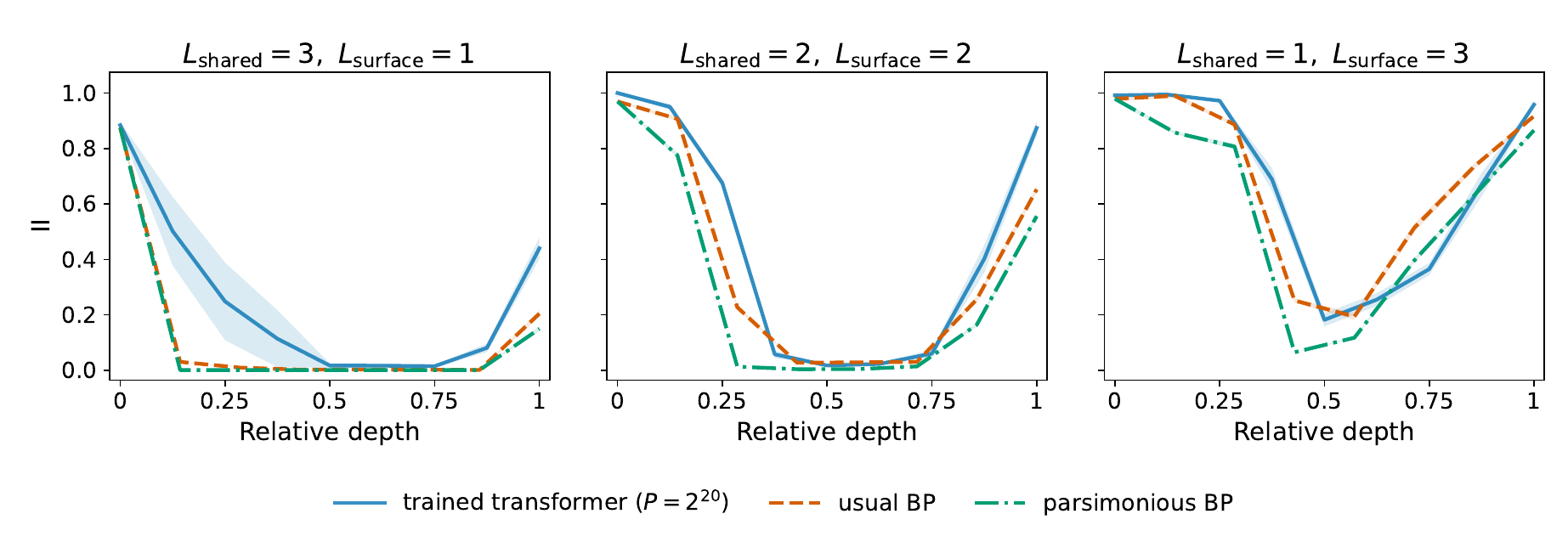}
    \caption{II for parsimonious implementation of BP where the only latent messages stored correspond to completely observed subtrees -- compared with the ordinary BP and trained transformers. The qualitative behaviour is unaffected.}
    \label{figapp:II_parsimonious}
\end{figure}

Finally, a parsimonious implementation encodes a level-$\ell$ upward message
only after its subtree is complete, i.e. when $t$ is divisible by $s^\ell$.
The construction used in the main text also represents partial-subtree
messages by marginalizing unobserved leaves. This choice does not
qualitatively affect the results as we show in Fig.~\ref{figapp:II_parsimonious}.

\paragraph{Depth-limited inference.}
With finite training data, transformers recover the production rules
progressively from the surface upward \citep{cagnetta2024towards}. We model a
network that has learned the rules only through level $\ell_\ast<L$ by
replacing every unresolved factor with a uniform factor over all possible
productions:
\begin{equation}
\label{eq:bp_unknown_factor}
\psi_\ell^{\mathrm{unknown}}(b,\ba)
=
1,
\qquad
b\in\cA,\quad \ba\in\cA^s,\quad \ell>\ell_\ast.
\end{equation}
Thus, every child tuple is considered equally possible, including tuples that
are invalid under the true grammar. 

Consequently,
\begin{equation}
\label{eq:bp_uniform_unknown_messages}
\left[\bu_t^{(\ell)}\right]_b
=
\frac{1}{v},
\qquad \ell>\ell_\ast,
\qquad\text{and}\qquad
\left[\bd_t^{(\ell_\ast)}\right]_b
=
\frac{1}{v}.
\end{equation}
The uniform message at level $\ell_\ast$ provides the boundary condition for
the downward recursion, which then proceeds through the learned factors at
levels $\ell_\ast,\ldots,1$.

Unlike a truncated computation, this construction retains the complete
$(2L{-}1)$-step BP computation. Updates involving unresolved levels carry only
uniform, sample-independent messages and therefore introduce no additional
information into the representations.

When $\ell_\ast<L_{\mathrm{surface}}$, only language-specific rules have been
learned. Cross-lingual similarity emerges when $\ell_\ast$ reaches
$L_{\mathrm{surface}}$ and strengthens as further shared levels are recovered.
Distinct depths $\ell_\ast^A$ and $\ell_\ast^B$ model unequal exposure to the
two languages.

\subsection{encoder-only shortcut, i.e. one-step decoding.}
\label{subsecapp:encoding_only}
Fixed-topology grammars admit an $(L{+}1)$-step shortcut in which the upward
messages are combined directly to predict the next token, avoiding a
sequential downward pass \citep{zhou2026how}. We call this an
``encoder-only'' computation (referring to the computational algorithm, not the transformer architecture).

Depending on the training-set size and code-switching rate $\epsilon$, trained
transformers approach either this shortcut or the full encoder--decoder
computation. Code-switching requires the predictor to determine which
language-specific rules govern individual subtrees, favoring an explicit
downward computation; we study this transition in
Appendix~\ref{secapp:transition}. Moreover, the shortcut relies on knowing
the tree topology in advance. In
Appendix~\ref{secapp:varying_topology}, we show that varying-topology grammars
require the full encoder--decoder computation. Since natural language is
generally modeled using variable-topology parse trees, this motivates our
focus on $\epsilon=0.1$, for which trained transformers consistently exhibit
the encoder--decoder structure.

Here, we recall for completeness how  the transformer can implement an encoder-only computation, as discussed in \citep{zhou2026how}.
Indeed, encoder-only algorithm composes all downward updates into a single function evaluated by the final transformer block. After the encoding phase, the network has computed the messages from the sibling sub-trees that contain observed tokens. Denote these messages by
\[
u_{0,t},u_{1,t},\ldots,u_{L-1,t}.
\]
For a fixed tree topology, the subtree associated with each message and the position at which it is stored are deterministic functions of \(t\).

The usual BP decoder processes these messages sequentially. Schematically,
\[
d_{L-1,t}=f_{L-1}(u_{L-1,t}),\qquad
d_{\ell-1,t}=f_{\ell-1}(d_{\ell,t},u_{\ell-1,t}),
\]
where \(d_{0,t}=p(x_{t+1}\mid x_{1:t})\). Substituting each update into the next eliminates the intermediate downward messages and gives
\[
d_{0,t}
=
F(u_{0,t},u_{1,t},\ldots,u_{L-1,t})
\]
for a fixed function \(F\). Thus, sequential decoding and one-step decoding differ only in how the same function is evaluated.  \citet{zhou2026how} provides an explicit construction showing that a single transformer block can perform this computation.

\subsection{Transformers approximate BP computation}
\label{subsecapp:layerwise_prediction}

\begin{figure}[!h]
    \centering
    \includegraphics[width=\linewidth]{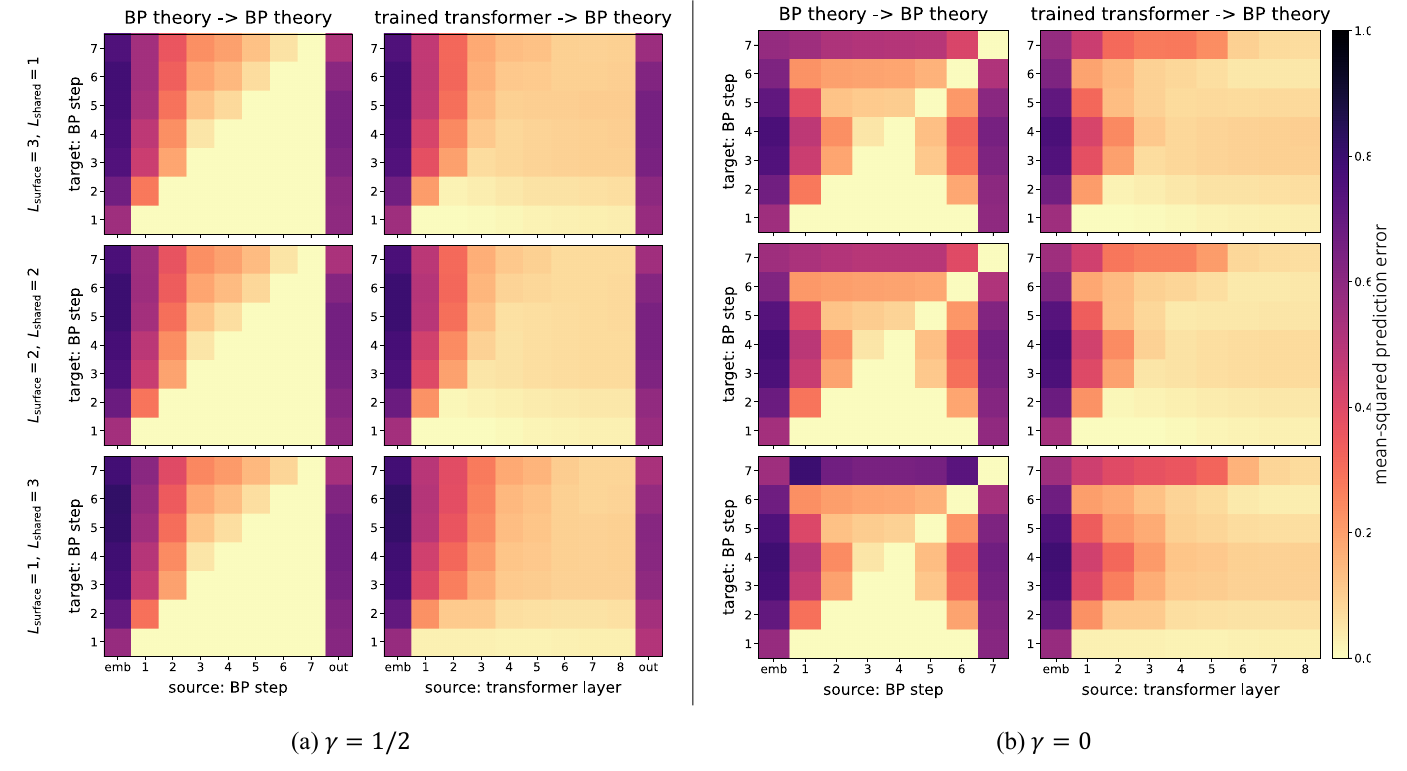}
    \caption{Mean-squared error when predicting layerwise representations from one network using another, for $\gamma=1/2$ \textbf{(a)} and $\gamma=0$. \textbf{(b)} BP encodings. Transformer $\rightarrow$ theory predictions confirm the expected layerwise encoding of BP messages in trained transformers. These heatmaps also match with theory $\rightarrow$ theory predictions, further solidifying the correspondence. The match is better for $\gamma=1/2$.}
    \label{figapp:layerwise_prediction}
\end{figure}

Here we provide direct evidence that transformers trained on MRHM data approximately implement the BP algorithm described above. We do so by linearly predicting the representations of the theoretical network from transformer representations. Figure~\ref{figapp:layerwise_prediction} shows a heatmap of mean-squared error predicting each layer $\ell$ of the theoretical network from each transformer layer $l$. Indeed, the transformer layers accurately predict theoretical representations corresponding to BP messages it is expected to encode. We compare this heatmap with a prediction from one theoretical model to another (this is trivial on the diagonal, but nontrivial elsewhere). The two prediction heatmaps, transformer $\rightarrow$ theory and theory $\rightarrow$ theory show similar layerwise behaviour, further strengthening the correspondence. Note that the correspondence is especially clear for the  encoding of BP where used messages decay at a rate $\gamma=1/2$ compared to $\gamma=0$ where messages are immediately eliminated, as shown in panel (b) of Figure~\ref{figapp:layerwise_prediction}.

\section{Training details}
\label{secapp:training_details}

We trained 8-layer decoder-only transformers with embedding dimension 512, four attention heads, and ReLU activations using next-token cross-entropy loss. Optimization used AdamW (\(\beta_1=0.9,\beta_2=0.999\)) with peak learning rate \(10^{-3}\), weight decay \(10^{-2}\), and batch size \(1024\). We used a warmup–stable–decay schedule: linear warmup over the first \(10\%\) of training, a constant learning rate for the middle \(80\%\), and cosine decay over the final \(10\%\) to \(10^{-2}\) times the peak rate. Models were trained for up to 128 epochs, and analyses used the checkpoint with the lowest next-token validation loss. Reported results are averaged over seeds \(42,420,4200,\) and \(42000\).
For testing, we use a held-out set of $2^{15}$ sentences. 

\begin{figure}[!h]
    \centering
    \includegraphics[width=\linewidth]{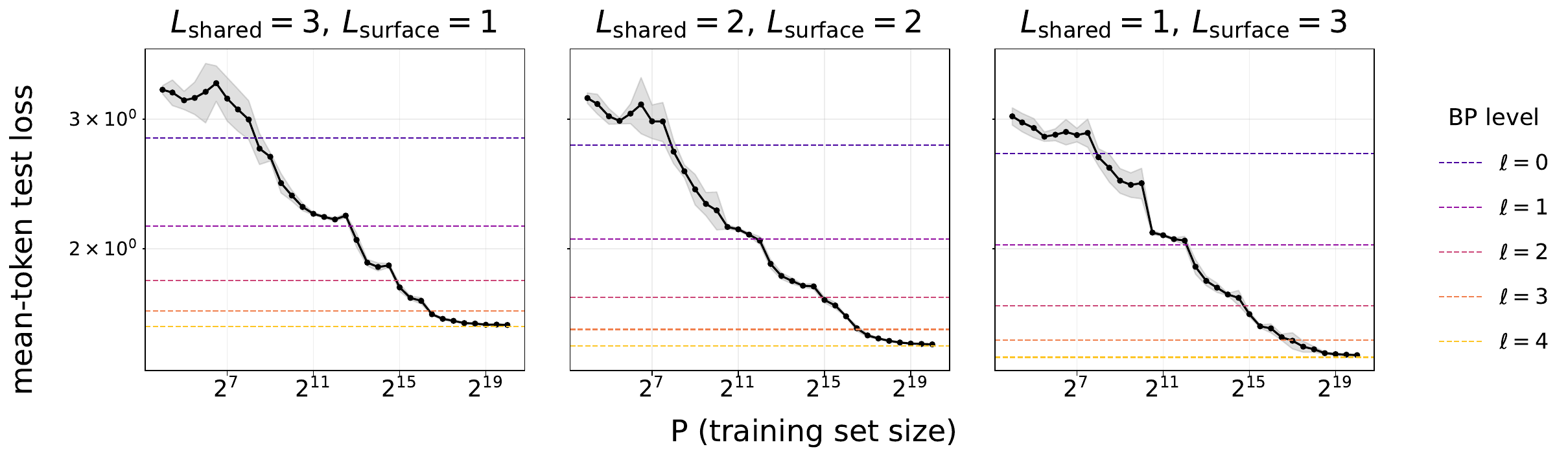}
    \caption{Test loss for transformers trained on MRHM data with $\epsilon=0.1$ and different training set sizes $P$. The loss scales with $P$, showing sequential plateaus at the theoretical entropies computed with depth-limited BP.}
    \label{figapp:loss_scaling}
\end{figure}

In Fig.~\ref{figapp:loss_scaling} we show the test loss of trained transformers as a function of the training set size $P$. The stepwise decrease in the loss follows theoretical entropies calculated with depth-limited BP. 

\section{Similarity measures}
\label{secapp:similarity_measures}

We measure representational similarity using information imbalance (II) and
mutual $k$-nearest-neighbor overlap (mKNN). These quantities capture neighbourhood similarity -- II looks at the nearest neighbour, mKNN compares a larger local neighborhood.

The ``network'' may be either a trained transformer or the theoretical network
constructed by encoding BP messages. We generate $N$ pairs of translated
sentences $\{(\bx_n^A,\bx_n^B)\}_{n=1}^N$ in languages $A$ and $B$. At layer
$l$, we extract the corresponding token-averaged representations
$\bh_n^{A,(l)}, \bh_n^{B,(l)} \in\bbR^{d_l}$, and stack them into matrices
\(
\bH^{A,(l)}, \bH^{A,(l)}\in\bbR^{N\times d_l},
\)

For II and mKNN measurements in the RHM setup, we set $N=2048$.

\paragraph{Information imbalance.}
Information imbalance asks whether the closest neighbor of each sentence in
one representation remains nearby in the other. For each sentence $n$, let
\[
j_A(n)
=
\argmin_{j\neq n}
d_A\!\left(\bh_n^{A,(l)},\bh_j^{A,(l)}\right)
\]
be its nearest neighbor in space $A$, and let $r_B(n,j)$ denote the rank of
sentence $j$ among the neighbors of $n$ in space $B$, with rank one denoting
the nearest neighbor. The directional information imbalance is
\[
\Delta(A\!\rightarrow\!B)
=
\frac{2}{N^2}
\sum_{n=1}^{N}r_B\!\left(n,j_A(n)\right).
\]
Thus, II first identifies a local relationship in $A$ and then measures how
well that relationship is preserved in $B$. If the two representations have
similar local geometry, $j_A(n)$ will also have a low rank in $B$, giving a
small value of $\Delta(A\!\rightarrow\!B)$. If their neighborhood structures
are independent, the expected rank is approximately $N/2$, and the
normalization gives $\Delta(A\!\rightarrow\!B)\approx1$.

II is directional: $\Delta(A\!\rightarrow\!B)$ need not equal
$\Delta(B\!\rightarrow\!A)$. For example, one representation may separate
sentences that are indistinguishable in the other. In the main figures, we
report the symmetrized value
\[
\overline{\Delta}(A,B)
=
\frac{1}{2}
\left[
\Delta(A\!\rightarrow\!B)+\Delta(B\!\rightarrow\!A)
\right].
\]

\paragraph{Mutual $k$-nearest-neighbor overlap.}
Whereas II follows only the closest neighbor, mKNN compares the full set of
the $k$ nearest neighbors of each sentence. Let
$\mathcal N_A^k(n)$ and $\mathcal N_B^k(n)$ denote these sets in the two
representation spaces. We define
\[
\operatorname{mKNN}_k(A,B)
=
\frac{1}{N}
\sum_{n=1}^{N}
\frac{
\left|\mathcal N_A^k(n)\cap\mathcal N_B^k(n)\right|
}{k}.
\]
The score lies between 0 and 1. A value of 1 means that every sentence
has exactly the same $k$ nearest neighbors in the two spaces, while smaller
values indicate less agreement between their local geometries. Unlike II,
mKNN is symmetric by construction.

We use $k=10$ and identify neighbors using cosine similarity after
$\ell_2$-normalizing each representation. Exact distance ties occur more
frequently in the discrete BP representations than in the trained networks.
We therefore compute the expected overlap under independent uniform tie
breaking, avoiding artificial agreement caused by deterministic,
index-based tie breaking.

The scale probed by mKNN depends on the choice of $k$. Moreover, for
independent neighborhood structures its expected value is approximately
\(
\frac{k}{N-1}.
\)
Thus, unlike the normalized II score, raw mKNN values depend explicitly on
both $k$ and the evaluation-set size $N$. Comparisons across training-set
sizes $P$ are valid in our experiments because we hold $N$ and $k$ fixed,
but mKNN is not an intrinsic, parameter-free measure of neighborhood
agreement in the same sense as II.

\section{Algorithmic transition with $\epsilon$ and $P$}
\label{secapp:transition}

\begin{figure}[!h]
    \centering
    \begin{subfigure}{0.8\linewidth}
        \centering
        \includegraphics[width=\linewidth]{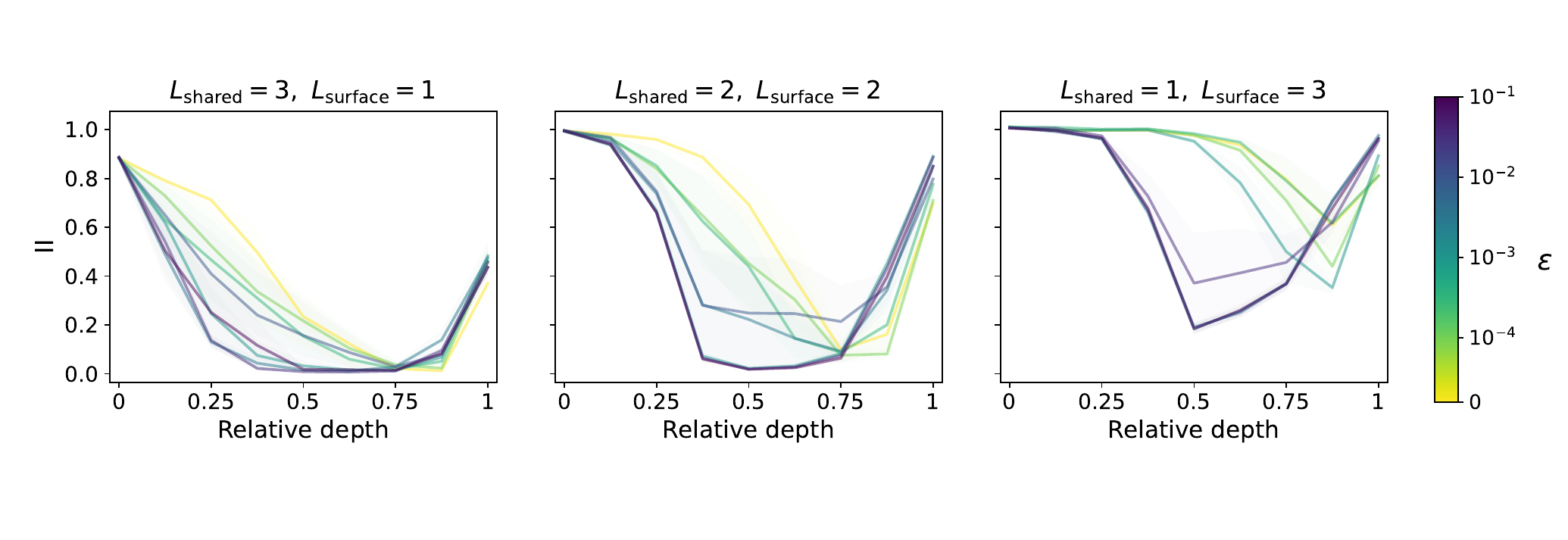}
        \caption{II profiles for various $\epsilon$, for fixed $P=2^{20}$. encoder-only to encoding-decoding transition can be seen with increasing $\epsilon$.}
    \end{subfigure}
    \begin{subfigure}{0.8\linewidth}
    \centering
        \includegraphics[width=\linewidth]{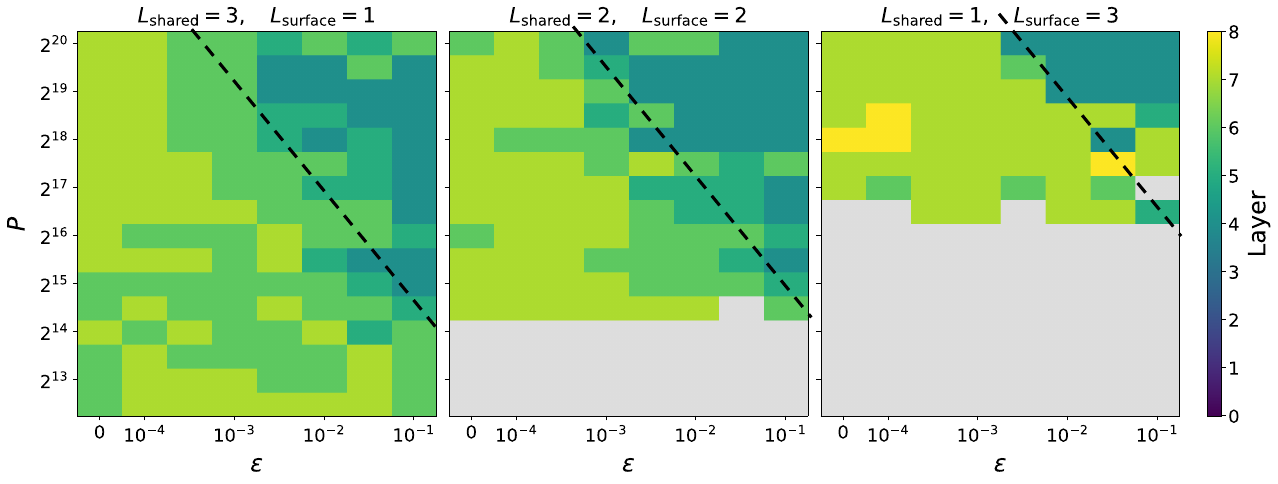}
        \caption{Tracking the transformer layer with the lowest II shows the algorithmic transition wrt $\epsilon$ and $P$. Each grid-box is a separately trained transformer. Cases where minimum II was greater than 0.8 are grayed out. Comparing this with Fig.~\ref{fig:latent_probing_epsilon} in the main text, we find that the algorithmic transition approximately co-occurs with coordinate alignment.}
    \end{subfigure}
    \caption{Transition from encoder-only to encoding-decoding computations in transformers trained on MRHM data with various $\epsilon$ and $P$. Interestingly, the algorithmic transition co-occurs with coordinate alignment (Fig.~\ref{fig:latent_probing_epsilon}).}
    \label{figapp:transition}
\end{figure}

In trained transformers the transition between encoding-decoding and encoder-only behaviour is governed by two factors: the code-switching rate $\epsilon$ and training set size $P$. One way to study this transition is to train transformers for many different $P$ and $\epsilon$ values, and identify the layer that shows the lowest II, and therefore the highest similarity. The Encoder-decoder design predicts highest similarity in the middle layers, whereas encoder-only design predicts that at the penultimate layer. Indeed, in Fig.~\ref{figapp:transition} we show that this observable can identify the algorithmic transition with $\epsilon$ and $P$.

\begin{figure}[!h]
    \centering
    \includegraphics[width=\linewidth]{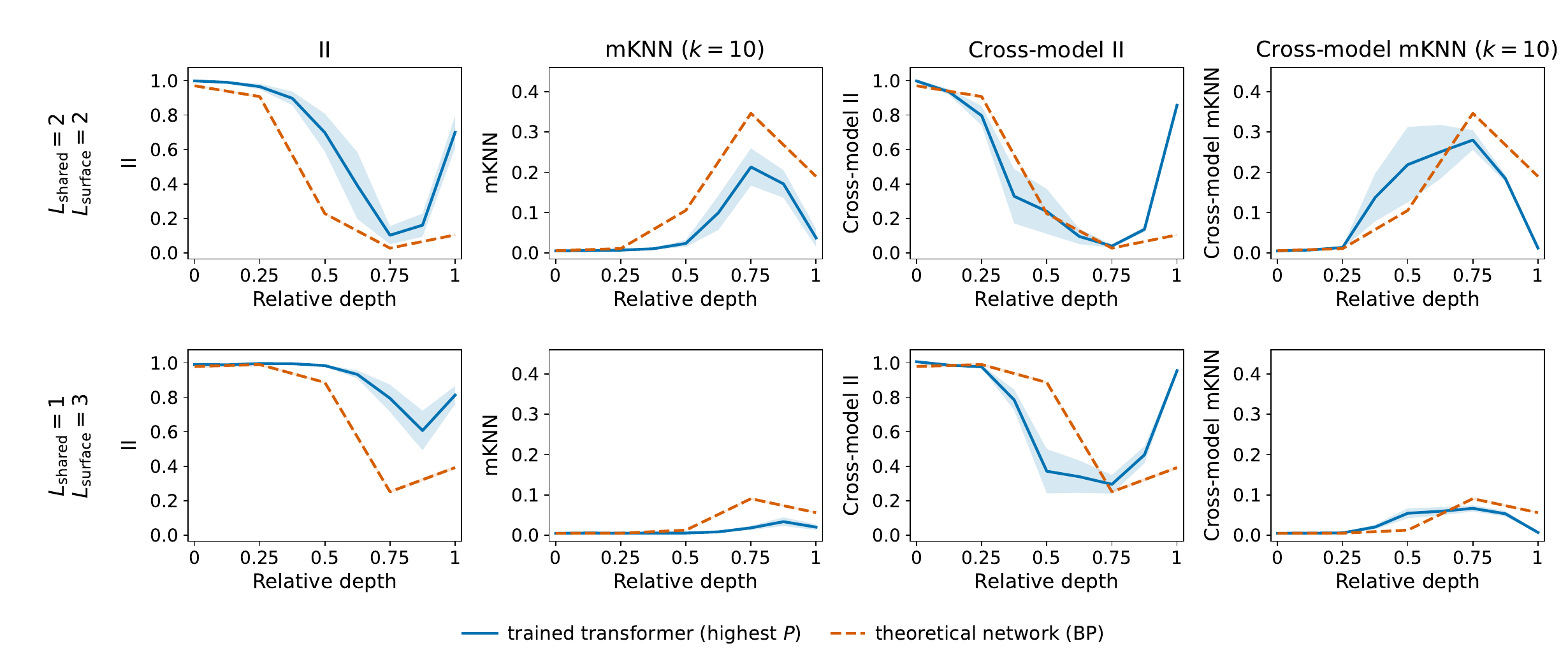}
    \caption{$\epsilon=0$ version of Fig.~\ref{fig:similarity} in the main text. In this case, the layerwise similarity trends in trained transformers are better captured by that of a theoretical network constructed with an $(L{+}1)$-step encoder-only BP algorithm.}
    \label{figapp:similarity_encoding_only}
\end{figure}

\section{Varying the Number of Transformer Layers}
\label{secapp:varying_n_layers}

\begin{figure}[!h]
    \centering
    \includegraphics[width=0.8\linewidth]{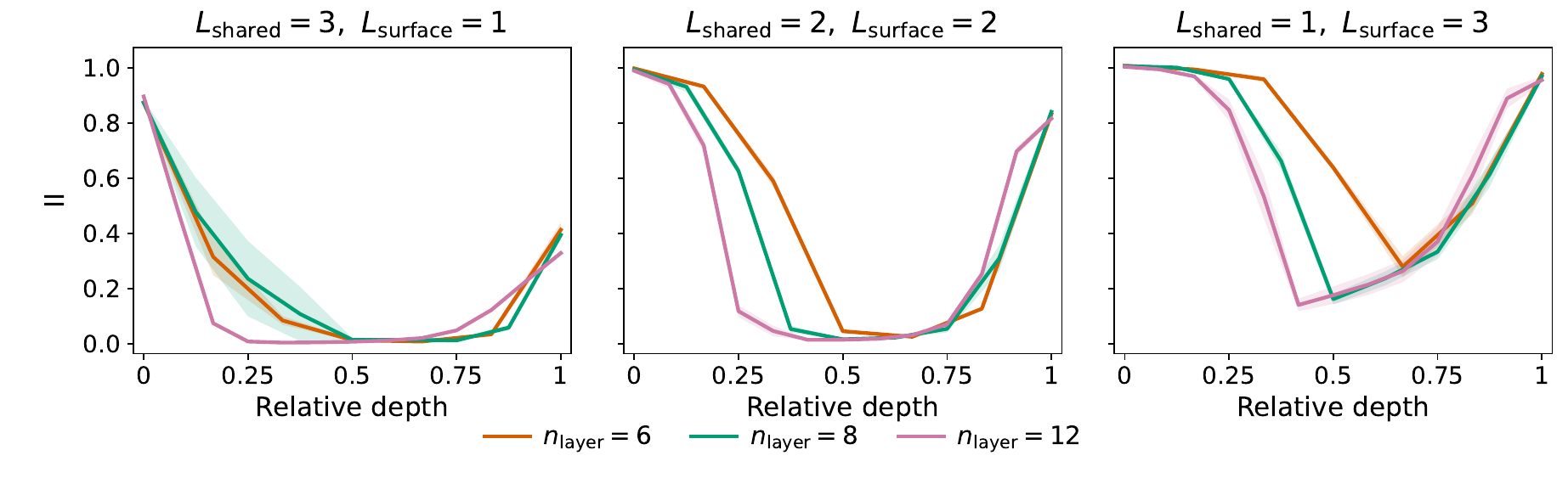}
    \caption{Layerwise similarity profiles with varying number of transformer layers. The trend remains qualitatively unchanged, evidencing that transformers stretch the BP computation steps among its layers when more depth is available.}
    \label{figapp:varying_n_layers}
\end{figure}

The correspondence between trained transformers and theoretical networks constructed with BP does not require identical number of layers between the two networks. This is already clear since all experiments in the main text use an 8-layer transformer, while the theoretical network has 7 layers, corresponding to the $2L-1 = 7$ steps of the encoding-decoding BP computation. Transformers tend to ``stretch" the same computation over all available layers, resulting in dilation of the expected layerwise trends. We explicitly test this by training transformers of varying depths on the same MRHM data. As shown in Fig.~\ref{figapp:varying_n_layers}, the layerwise similarity trends remain qualitatively unchanged with varying transformer depth. Note that a minimum transformer depth is still required for the computation, below which the performance would suffer from suboptimal predictions.

\section{Varying topology}
\label{secapp:varying_topology}

We study the varying tree topology RHM \citep{parley2026deep} and extend it to MRHM.
 
\paragraph{Varying-tree MRHM.}
We keep the notation of Section~\ref{sec:multilingual_rhm}, but each node now has two or three children. At each level, every symbol $b\in\mathcal{A}$ has $m_2$ binary rules $b\to(a_1,a_2)$ and $m_3$ ternary rules $b\to(a_1,a_2,a_3)$, whose child tuples are sampled without replacement from $\mathcal{A}^2$ and $\mathcal{A}^3$, so each tuple has a unique parent.
Each node draws an arity $s\in\{2,3\}$ with probability $p_s$, then one of its $m_s$ $s$-ary rules uniformly; $p_2\in\{0,1\}$ recovers the fixed-tree RHM.
A level-$\ell$ node thus spans between $2^\ell$ and $3^\ell$ tokens, and sentence lengths range over $[2^L,3^L]$, so sentence length varies and the segmentation of the token string into subtrees is not known in advance: the
tree topology must be inferred along with the symbols.
 Ambiguity is controlled by the fractions of admissible tuples, $f_2=m_2/v$ and $f_3=m_3/v^2$.
The multilingual version shares the rules above $L_{\text{surface}}$. A translation pair shares its upper tree, including its topology, while its lower subtrees differ.

\begin{figure}[ht!]
    \centering
    \includegraphics[width=0.3\linewidth]{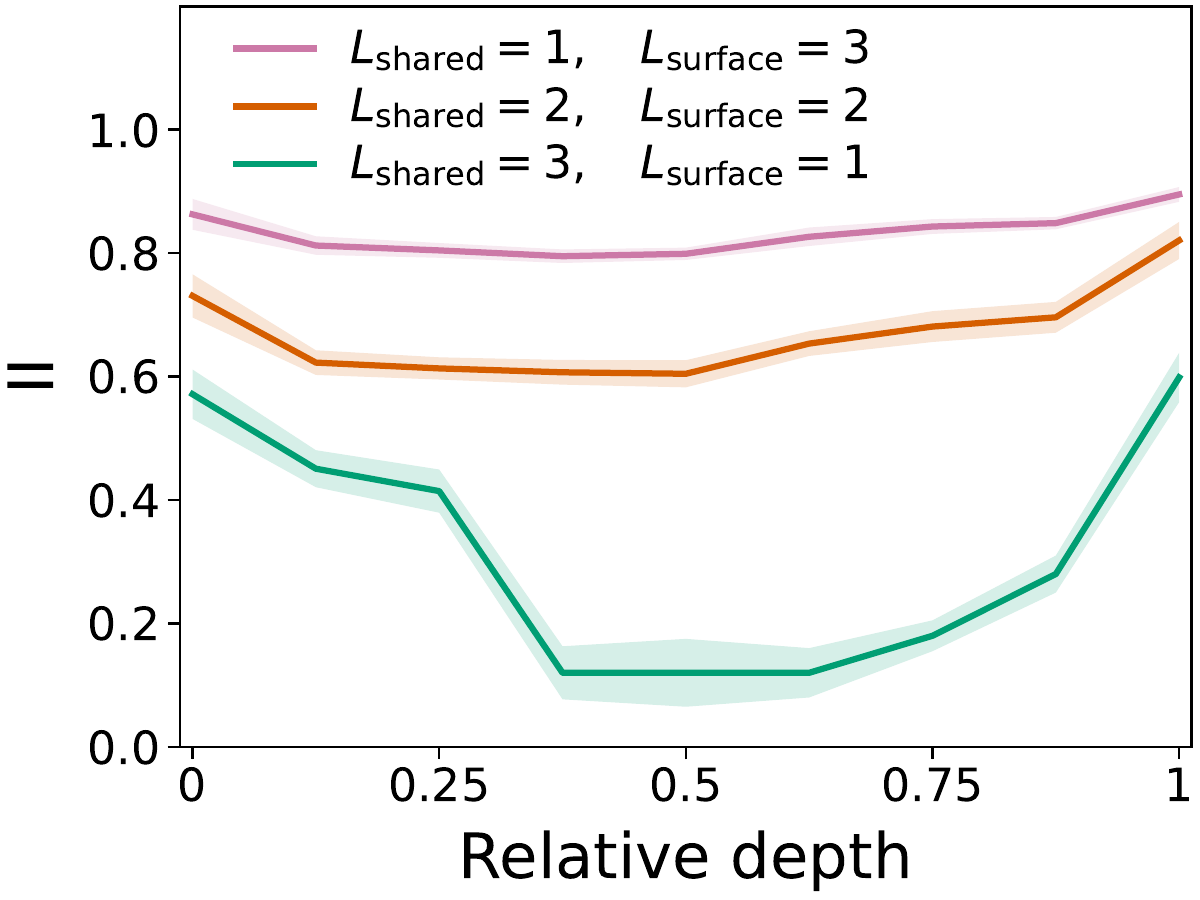}
    \caption{Cross-lingual similarity with varying tree topology. II is shown against relative transformer depth for three splits of a generative hierarchy with $L=4$. Across all three splits, II reaches an intermediate-layer minimum and rises toward the output, consistent with the encoder--decoder picture.}
    \label{fig:II_varying_top}
\end{figure}

\paragraph{Encoder--decoder structure with varying topology.}
Figure~\ref{fig:II_varying_top} shows that cross-lingual II reaches its minimum at intermediate transformer layers and increases again toward the output. This profile is consistent with an encoder--decoder computation, in which shared abstractions emerge during encoding, followed by decoding into language-specific predictions.

\section{Coordinate alignment}
\label{secapp:coordinate_alignment}

We train linear probes from transformer representations of  RHM sentences to the corresponding latent variables that generated the sentences. Each transformer layer ($l$), and each latent level ($\ell$) and position ($i$) has a separate probe, which linearly classifies the latent variable. The test accuracy of this probing tells us whether latents at a given RHM level are encoded in a given transformer layer. 
For probing a latent, we specifically focus on the representations at the token index $i$ that corresponds to the last token of the subtree spanned by that latent -- so that all the information about that latent is causally available to the transformer representation.

To measure coordinate alignment, we use \emph{cross-probing}: independent probes are trained on (pure) language-A and language-B sentences. We test language-A probes on language-B sentences and vice-versa. If we denote the trained linear probes, for latent level $\ell$ and position $i$, and transformer level $l$, by the weight \(\mathbf W^{A(\ell, l)}_i\) and bias \(b^{A(\ell, l)}_i\), then the cross-probing predictions can be written as 
\[
\Hat{y}_i^{A\rightarrow B(\ell,l)}
=
\argmax_z
\left[
\mathbf W^{A(\ell, l)}_i \bh^{B(l)}_i+\mathbf b^{A(\ell, l)}_i
\right]_z
\]

Since the latent variables above the split level $L_{\mathrm{surface}}$ are shared, the probes in both languages decode the same information. For cross-probing to be successful, this shared information should be represented in shared coordinates. 

Note that below $\ell = L_{\mathrm{surface}}$ coordinates never align, because the latent variables themselves are language-specific. Therefore, we report the cross-probing accuracy at $\ell = L_{\mathrm{surface}}$, which is the first shared layer. We find that once this level is aligned, the levels above also align.

The cross-probing results are shown in Fig.~\ref{fig:latent_probing_epsilon}, where we find that the coordinate alignment occurs with increasing code-switching rate $\epsilon$ and training set size $P$.


\section{Layerwise Novelty}
\label{secapp:novelty}

\paragraph{Definition.}
Let $h_{n,t}^{(k)}\in\mathbb{R}^{1\times d_k}$ denote the representation of token $t$ in sentence $n$ at layer $k$. We define layerwise novelty as the residual after subtracting an affine prediction based on the preceding layer:
\begin{equation}
    r_{n,t}^{(k)}
    =h_{n,t}^{(k)}-f_k\!\left(h_{n,t}^{(k-1)}\right),
    \qquad f_k(h)=hW_k+b_k.
\end{equation}
We average these residuals over valid tokens to obtain the sentence representation used for cross-lingual comparison.

\paragraph{Tokenwise affine fit.}
For each layer transition, we fit a single affine map shared across all token positions and sentences in the fitting set using ridge regression:
\begin{equation}
    (W_k,b_k)=\operatorname*{arg\,min}_{W,b}
    \left\{
    \frac{1}{S}\sum_{n=1}^{S}\frac{1}{T_n}\sum_{t=1}^{T_n}
    \left\|h_{n,t}^{(k)}-h_{n,t}^{(k-1)}W-b\right\|_2^2
    +\lambda\|W\|_F^2
    \right\}.
\end{equation}
Here $S$ is the number of fitting sentences, $T_n$ is the number of valid tokens in sentence $n$, and $\lambda$ is the regularization parameter, selected to minimize the affine map's mean squared prediction error on a validation set. The factor $1/T_n$ gives each sentence equal total weight.

\begin{figure}[!h]
    \centering
    \begin{subfigure}{\linewidth}
        \centering
        \includegraphics[width=\linewidth]{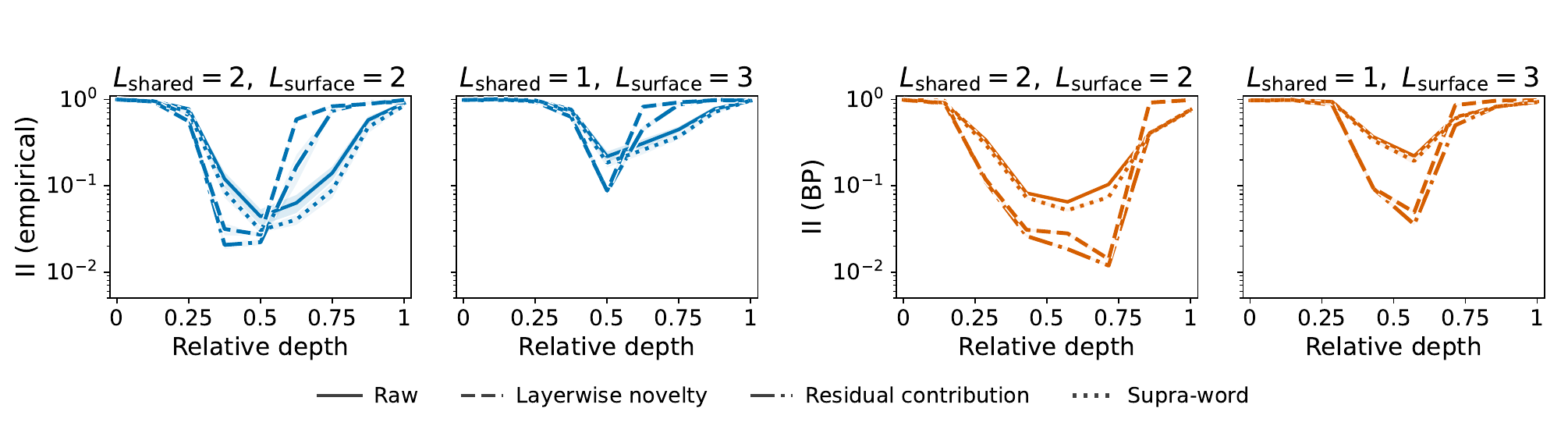}
        \caption{II}
    \end{subfigure}
    \begin{subfigure}{\linewidth}
        \centering
        \includegraphics[width=\linewidth]{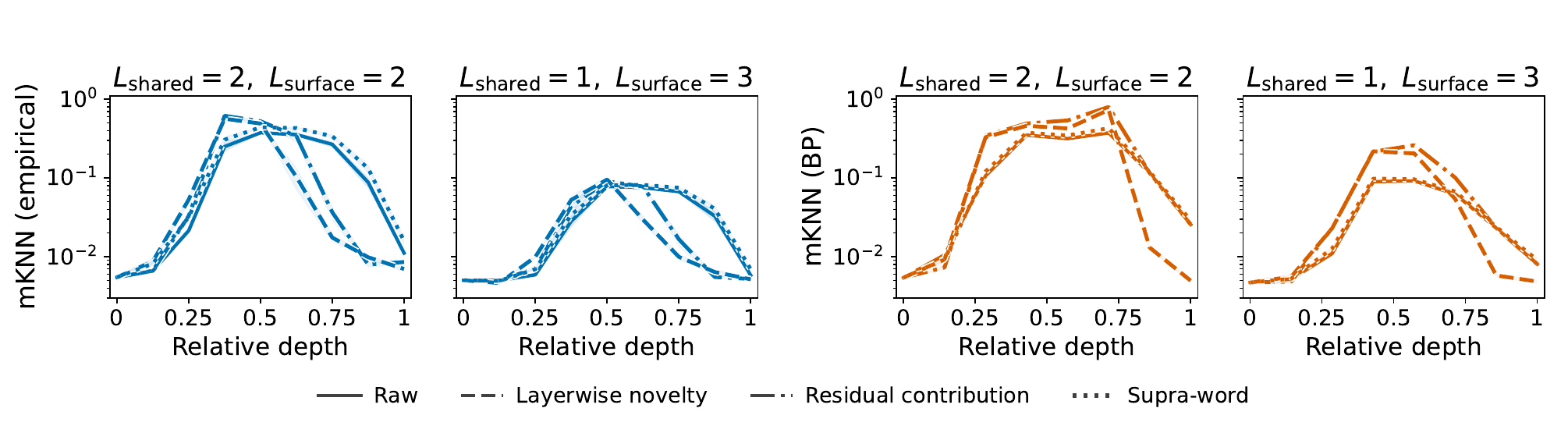}
        \caption{mKNN}
    \end{subfigure}
    \caption{Extended version of Fig.~\ref{fig:novelty}. Cross-lingual geometric similarity in trained transformers and theoretical BP networks is measured using II \textbf{(a)} and mKNN \textbf{(b)} for raw representations, layerwise novelty, the residual contribution, and the supra-word residual. Lower II and higher mKNN indicate greater similarity.}
    \label{figapp:rhm_novelty}
\end{figure}

Figure~\ref{figapp:rhm_novelty} extends the RHM analysis in Fig.~\ref{fig:novelty} by reporting both II and mKNN for trained transformers and theoretical BP networks, with the residual contribution $h^{(k)}-h^{(k-1)}$ and the supra-word residual as additional baselines. Both metrics support the same qualitative picture: novelty enhances cross-lingual similarity during encoding but reduces it during decoding relative to raw representations, consistent with the removal of retained shared information. 

\begin{figure}[!h]
    \centering
    \begin{subfigure}{\linewidth}
        \centering
        \includegraphics[width=\linewidth]{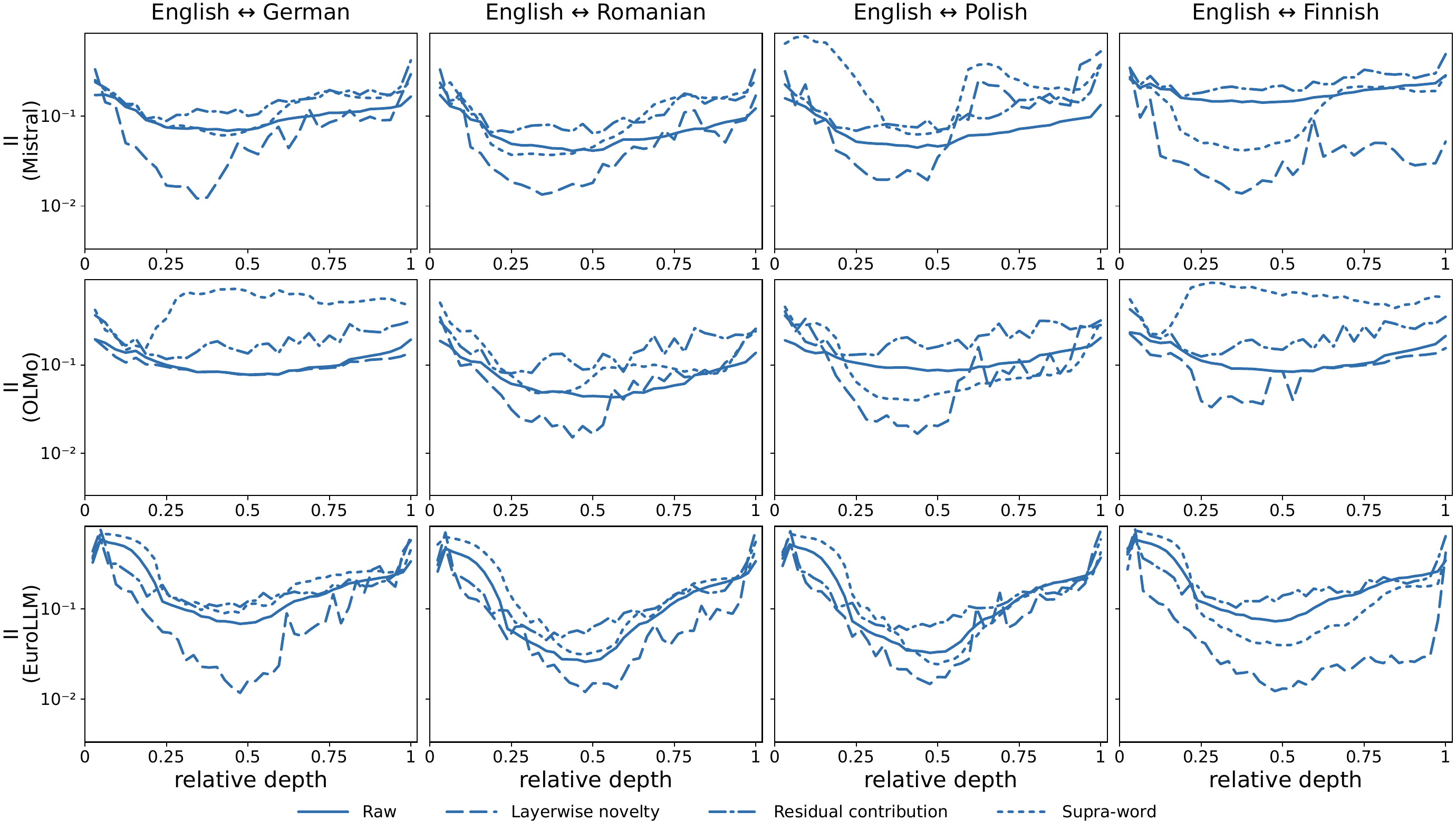}
        \caption{II}
    \end{subfigure}
    \begin{subfigure}{\linewidth}
        \centering
        \includegraphics[width=\linewidth]{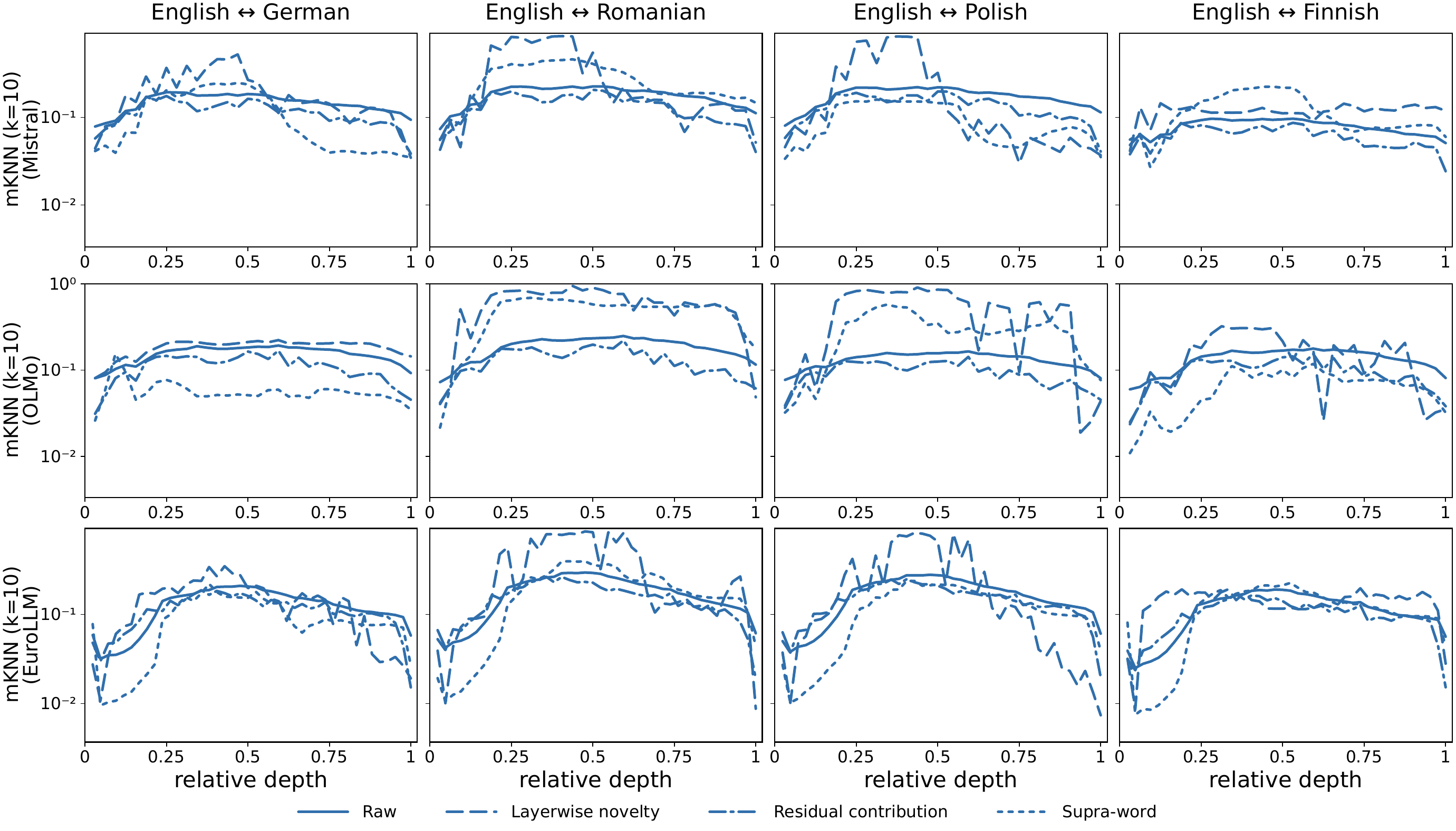}
        \caption{mKNN}
    \end{subfigure}
    \caption{Extended version of Fig.~\ref{fig:novelty_natural_language}. Cross-lingual geometric similarity in Mistral-7B-v0.1, OLMo-3-7B, and EuroLLM-9B is evaluated on parallel English--German, English--Romanian, English--Polish, and English--Finnish sentences from Europarl. II \textbf{(a)} and mKNN \textbf{(b)} are measured for raw representations, layerwise novelty, the residual contribution, and the supra-word residual. Lower II and higher mKNN indicate greater similarity.}
    \label{fig:ii_mknn_natural_language}
\end{figure}

Figure~\ref{fig:ii_mknn_natural_language} extends the natural-language analysis to Mistral-7B-v0.1, OLMo-3-7B, and EuroLLM-9B \citep{martins2025eurollm9btechnicalreport} across four language pairs. Layerwise novelty generally yields lower II and higher mKNN than raw representations at intermediate layers, whereas improvements from the residual contribution and supra-word residual are less consistent across models and language pairs.

\end{document}